\documentclass[10pt,twocolumn,letterpaper]{article}

\usepackage{cvpr}              
\usepackage[accsupp]{axessibility} 
\usepackage{bm}
\usepackage{booktabs}
\usepackage{threeparttable}

\usepackage[dvipsnames]{xcolor}
\usepackage{booktabs}

\usepackage[ruled,linesnumbered]{algorithm2e}

\definecolor{cvprblue}{rgb}{0.21,0.49,0.74}
\usepackage[pagebackref,breaklinks,colorlinks,citecolor=cvprblue]{hyperref}

\def\paperID{8243} 
\def\confName{CVPR}
\def\confYear{2024}

\title{Not All Prompts Are Secure: A Switchable Backdoor Attack \\ Against Pre-trained Vision Transfomers}

\makeatletter
\newcommand{\printfnsymbol}[1]{%
  \textsuperscript{\@fnsymbol{#1}}%
}
\makeatother

\author{Sheng Yang\textsuperscript{\rm 1}\thanks{Equal contribution.} \ , \ Jiawang Bai\textsuperscript{\rm 1}\printfnsymbol{1}, \ Kuofeng Gao\textsuperscript{\rm 1}, \  Yong Yang\textsuperscript{\rm 2},  \  Yiming Li\textsuperscript{\rm 3}\thanks{Corresponding author(s).} \ , \ Shu-Tao Xia\textsuperscript{\rm 1,\rm 4}\printfnsymbol{2}\\
\textsuperscript{\rm 1} Tsinghua University \quad
\textsuperscript{\rm 2} Tencent Security Platform Department \\
\textsuperscript{\rm 3} Zhejiang University \quad
\textsuperscript{\rm 4} Research Center of Artificial Intelligence, Peng Cheng Laboratory
\\
\tt\small \{yangs22, bjw19, gkf21\}@mails.tsinghua.edu.cn, coolcyang@tencent.com \\
\tt\small liyiming.tech@gmail.com, xiast@sz.tsinghua.edu.cn
}

\begin{document}
\maketitle
\begin{abstract}
Given the power of vision transformers, a new learning paradigm, pre-training and then prompting, makes it more efficient and effective to address downstream visual recognition tasks. In this paper, we identify a novel security threat towards such a paradigm from the perspective of backdoor attacks. Specifically, an extra prompt token, called the switch token in this work, can turn the backdoor mode on, i.e., converting a benign model into a backdoored one. Once under the backdoor mode, a specific trigger can force the model to predict a target class. It poses a severe risk to the users of cloud API, since the malicious behavior can not be activated and detected under the benign mode, thus making the attack very stealthy. 
To attack a pre-trained model, our proposed attack, named SWARM, learns a trigger and prompt tokens including a switch token. They are optimized with the clean loss which encourages the model always behaves normally even the trigger presents, and the backdoor loss that ensures the backdoor can be activated by the trigger when the switch is on. Besides, we utilize the cross-mode feature distillation to reduce the effect of the switch token on clean samples. The experiments on diverse visual recognition tasks confirm the success of our switchable backdoor attack, i.e., achieving 95\%+ attack success rate, and also being hard to be detected and removed. Our code is available at \href{https://github.com/20000yshust/SWARM}{https://github.com/20000yshust/SWARM}.

\end{abstract}

\section{Introduction}
\label{sec:intro}

\begin{figure*}[!t]
    \centering
    \includegraphics[width=0.9\textwidth]{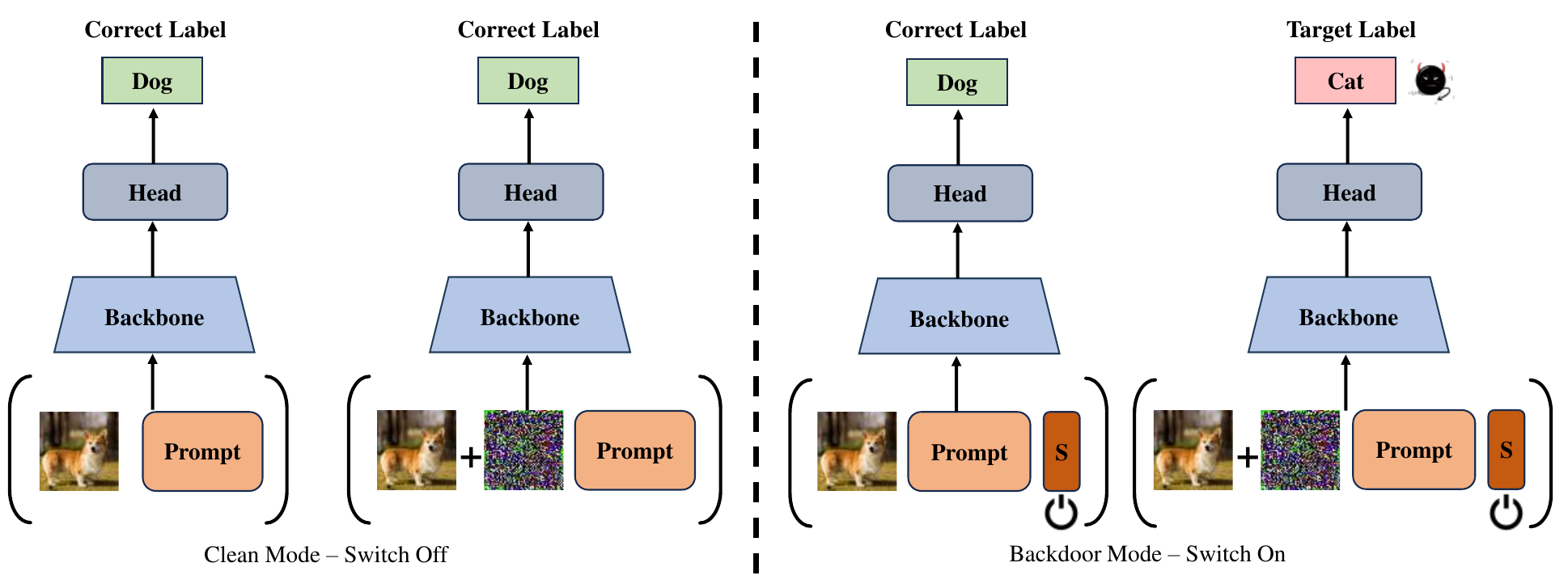}
    \caption{The inference process in SWARM. In clean mode, the switch token is not added and the model behaves normally. Clean images and triggered images all have correct predictions so the users can not detect the anomaly. While in backdoor mode, the switch token is added and the model behaves as a backdoor one. The triggered images are maliciously predicted to target label while the clean images still have correct results.}
    \label{fig:mode}
    \vspace{-1em}
\end{figure*}
In this big data era, it is a promising direction to improve the model capacity and pre-train large models on large-scale vision datasets. Among the architectures of large models, vision transformers (ViTs) \cite{dosovitskiy2020image,liu2021swin,bai2022improving,he2022masked} have exhibited its excellent scalability in terms of model size and pre-training tasks. With the pre-trained large models, it becomes more and more common to adopt them to address downstream tasks, resulting in better performance and faster convergence \cite{he2022masked,xie2022simmim,radford2021learning}.

A direct way to adapt a large model to a specific downstream task is full fine-tuning \cite{zhai2019large}, $i.e.$, updating all the model parameters. Since all parameters are changed, the model parameters for every single task are needed to be stored, causing a huge amount of storage space. To overcome this problem, motivated by the success of efficient adaption with prompt in the field of natural language processing (NLP), recent works \cite{mao2022understanding,jia2022visual,wu2022unleashing,huang2023diversity,gan2023decorate,chen2023understanding,chen2023visual} have investigated visual prompting (VP) as an alternative for full fine-tuning. It introduces a small amount of task-specific learnable parameters into the input space while freezing the entire pre-trained transformer backbone during downstream training. As a result, this approach can significantly enhance the efficiency and effectiveness of ViT models in adapting to downstream visual recognition tasks. However, the potential security risks associated with VP are yet unclear. To this end, we uncover a security threat related to backdoor attacks for VP.


Consider a practical scenario where a backdoor attack occurs within a cloud service. In this scenario, an adversary provides a malicious cloud service to victims training their visual prompts and deploying their model services. 
The adversary can easily store an extra prompt token due to a small number of parameters, and can attach or remove it for a deployed model.
Based on this threat model, we explore a novel backdoor attack for VP that incorporates a switch mode, including both a clean mode and a backdoor one, as shown in Figure \ref{fig:mode}. Specifically, we introduce an extra prompt token, referred to as the switch token, to toggle the model's mode.
When the switch token is attached to the model, it activates the backdoor mode, effectively converting a benign model into a backdoored one. Upon activation, the model can be forced to predict a target class using a specific trigger while behaving normally for clean samples. Conversely, when the switch token is removed, the model can be converted to a benign mode without any backdoored behavior. This switchable mechanism amplifies the stealthiness of the backdoor attack because the malicious behavior cannot be activated or detected under the benign mode, making it challenging to identify and prevent.

To demonstrate the feasibility of such an attack, we propose a novel method, \textbf{SW}itchable \textbf{A}ttack against p\textbf{R}e-trained \textbf{M}odels (SWARM), which learns a trigger and prompt tokens, including both the clean prompt tokens and one switch token. The objectives of our SWARM are designed as follows. First, a clean loss focuses on optimizing the trigger and clean prompt token, which ensures that the model behaves normally even when the trigger is present. Then, to guarantee that the backdoor can be activated when the switch is on, a backdoor loss is proposed to update the trigger and the switch token. Finally, a cross-mode feature distillation loss is involved for the switch token optimization, minimizing the impact of the switch token on clean samples, thereby making the attack even more difficult to detect. Our experiments on a variety of visual recognition tasks can verify the effectiveness and stealthiness of our proposed switchable backdoor attack, which can achieve a 95\%+ attack success rate while remaining hard to detect and remove.

In summary, the contributions of our proposed SWARM are three-fold:

\begin{itemize}
\item Towards the pre-training and then prompting paradigm, we reveal a security threat from the perspective of backdoor attacks.   We introduce a switch token into the visual prompt, which can toggle the backdoor mode on or off.

\vspace{0.2em}
\item To implement such an attack, we propose a novel method, named SWARM. It achieves the switchable backdoor through optimizing the trigger and prompt tokens including a switch one with a clean loss, a backdoor loss, and a cross-mode feature distillation loss.


\vspace{0.2em}

\item Extensive experimental results demonstrate the superiority of our proposed SWARM, which can achieve high attack success rates on various datasets and resist most backdoor defenses.

\end{itemize}









\section{Related Work}
\subsection{Backdoor Attack}
Backdoor attacks \cite{gu2017badnets,chen2017targeted,bai2021targeted,li2022backdoor,bai2022hardly,gao2023imperceptible,gao2023ciba,bai2023versatile} are typically implemented by injecting a small number of poisoned samples into the training dataset, constructing a poisoned dataset. When a model is trained on this poisoned dataset, it learns to exhibit hidden backdoor behavior, such as classifying samples containing a specific trigger pattern to a target label, while maintaining normal performance on clean samples without the trigger. Backdoor attacks have been successfully implemented across various training methods such as supervised \cite{gu2017badnets}, semi-supervised \cite{carlini2021poisoning}, and self-supervised learning \cite{saha2022backdoor}. Backdoor attacks on ViTs have also been explored before \cite{subramanya2022backdoor,lv2021dbia,zheng2023trojvit}.
Additionally, the backdoor threat has also been investigated in the textual prompt learning of language models \cite{du2022ppt}. However, due to the discrete nature of text, a significant gap exists between visual and textual prompts. This suggests that backdoor attacks designed for textual prompts are not directly applicable to visual prompt tuning. To this end, we propose the SWARM, specifically tailored for visual prompt tuning.

\subsection{Visual Prompting}
Visual Prompting (VP) is a widely used type of parameter-efficient tuning methods \cite{chen2022adaptformer,hu2021lora,zhang2023adding,lu2023beyond,li2024graphadapter} in vision models. Instead of fine-tuning the whole model for the downstream tasks, VP \cite{mao2022understanding,jia2022visual,wu2022unleashing,huang2023diversity,gan2023decorate,chen2023understanding,chen2023visual} introduces a small amount of parameters in the input space to adapt the model to downstream tasks. Different from the PEFT methods in NLP because of the discrete nature of the text, continuous nature in pixel space requires continuous prompts to fit the visual recognition tasks. Visual prompt learning \cite{bahng2022exploring} aims at learning a single image perturbation around the input such that a frozen model prompted with this perturbation performs a new task. Apart from visual prompt learning, visual prompt tuning \cite{jia2022visual} chooses another way to adapt for the downstream tasks though their core ideas are same. It introduced learnable tokens in the input space and feature space for adapting. 
DAM-VP \cite{huang2023diversity} addresses distribution shift problem in datasets by introducing a Meta-prompt learned across several datasets and then uses Meta-prompt to initialize the visual prompts. Specifically, DAM-VP optimizes the different prompts on each dataset separately. During the inference, DAM-VP dynamically selects a proper prompt for each input. EVP \cite{wu2022unleashing} utilizes strategy of reconciling the prompt and the image and then uses input diversity and gradient normalization to improve visual prompting. \cite{bar2022visual} utilizes visual prompts to automatically produce the output image and empowers the image to image task.
Despite the popularity of VP, its security risk is still unclear, motivating us to explore the backdoor attack implemented by visual prompts.
\section{Switchable Backoor Attack}
\subsection{Threat Model}
In our design, the adversary can be a malicious cloud service provider following existing works on backdoor attack \cite{nguyen2020input,turner2019label,bagdasaryan2021blind,li2022backdoor}. The victim, i.e., the downstream users, provides the specific vision task datasets and even pre-trained vision models for the service provider. Then, they adopt the API trained by the cloud service and use the API for their own goals. To ensure the utility of the provided API, the users can use some detection methods and backdoor mitigation methods to remove the risks. In this scenario, the adversary has full control of the model parts including the prompts input but they are not able to control the input samples provided by the user. Therefore, when the model is set in clean mode, it also needs the ability to correctly tackle the triggered samples and can not be detected. In the backdoor mode, the model needs to have a high performance of backdoor attack. Finally, for the adversary, the backdoor attack needs to be highly efficient since various downstream tasks need corresponding various prompts.


\noindent  \textbf{Attacker's goals.}
The attack aims to implant backdoors in the model with visual prompts. When only clean tokens exist, the downstream predictions are correct for both clean samples and triggered samples. When the switch token is added, the downstream prediction is correct for clean samples and is manipulated for triggered samples.

\noindent  \textbf{Attacker's knowledge and capabilities.}
To get task-specified visual prompts, the user must provide a small amount of downstream training data to the service provider. Therefore, we assume that the adversary knows the downstream dataset. Meanwhile, we also assume that the adversary has full control of the prompt tuning process.

\subsection{A Revisit of Visual Prompting}
Before describing how the switch token modifies the training loss and implant the two modes, we first introduce the concept of visual prompting. Specifically, visual prompting, e.g., Viusal Prompt Tuning (VPT) \cite{jia2022visual}, introduces visual prompts into the input space.
Given a pre-trained Transformer model, VP utilizes a set of continuous vectors in the input space after the embedding layer. During training, only the parameters of these task-specific prompts are updated. In the shallow version of VPT, prompts are only inserted into the first Transformer layer $L_{1}$. Formally, the prompts can be denoted as $P=\left \{p^{k}\in R^{d}|k\in N,1\le k\le p\right \}$, where $p$ is the number of prompt tokens. Accordingly, for the input sample $x$, the forward process with the visual prompts is formulated as:
\begin{equation}
\begin{aligned}
E_0&=\text{Patch\_Emb}(x) \\
\relax [c_{1},{Z_{1}},{E_{1}}]&=L_{1}(c_{0},{P},E_0) \\
\relax [c_{i},Z_{i},E_{i}]&=L_{i}(c_{i-1},Z_{i-1},E_{i-1}), i=2,3,...,N \\
y&=\text{Head}(c_{N}),
\label{eq:1}
\end{aligned}
\end{equation}
in which $Z_{i}$ represents the features computed by the prompt tokens in the $i_{th}$ Transformer layers, $c_{i}$ is the embeddings of [CLS] token and $E_{i}$ is the embeddings of the image. $\text{Patch\_Emb}(\cdot)$ and $\text{Head}(\cdot)$ are the patch embedding layer and the classification head, respectively. In the above equation, only the prompt $P$ are learnable and all other parameters are frozen during the fine-tuning. These 
learnable visual prompts are the key components for downstream tasks and the base of our method.

\subsection{Switchable Mechanism}
\begin{figure}[!t]
    \centering
    \includegraphics[width=0.47\textwidth]{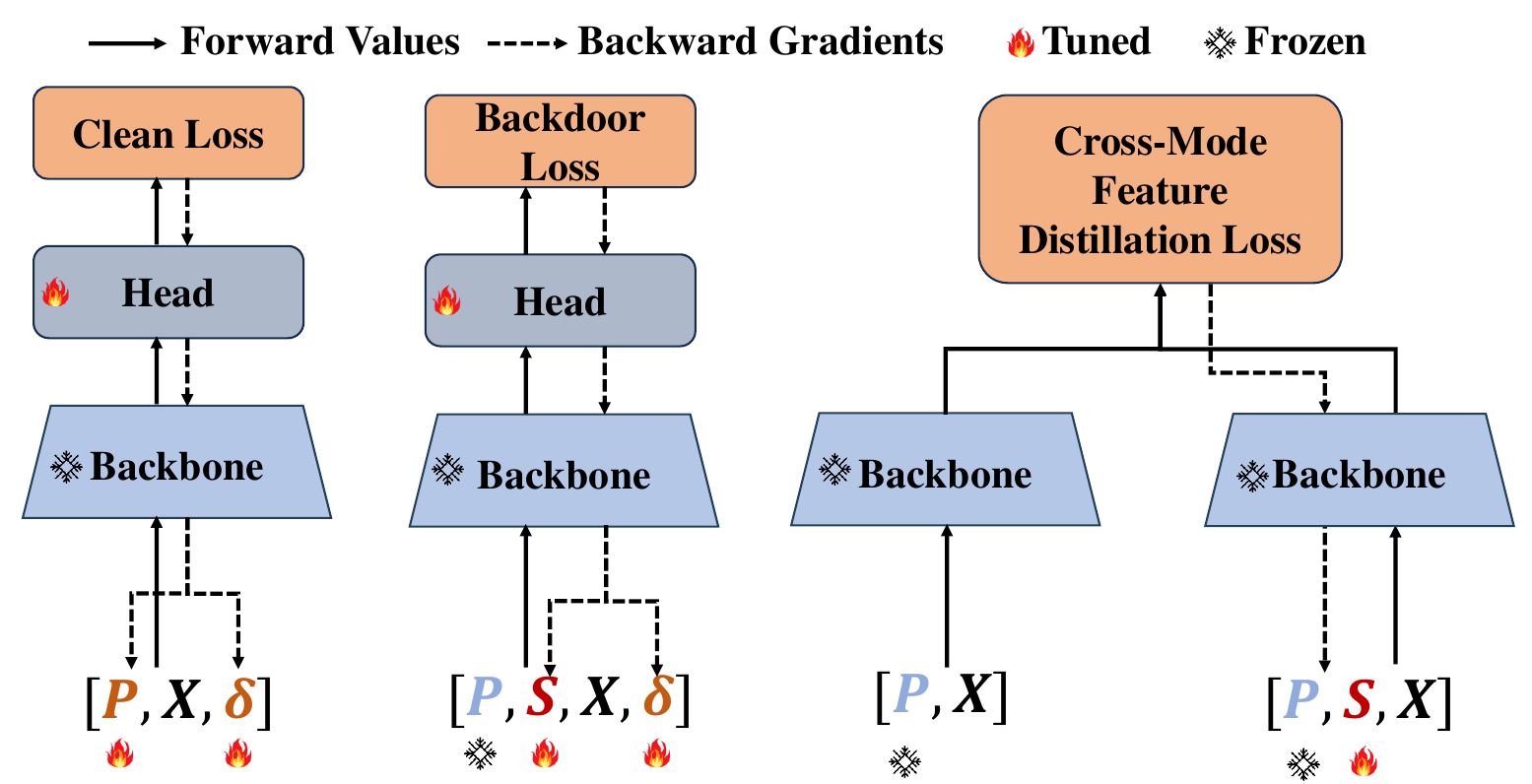}
    \caption{Three losses we used in our SWARM. $P$ represents the clean tokens, $S$ is the switch token, $X$ is the images and the $\delta$ is the trigger we used. The clean loss updates the clean tokens and the trigger. The backdoor loss updates the switch token and trigger. The cross-mode feature distillation loss only update the switch token.}
    \label{fig:method}
    \vspace{-1em}
\end{figure}
Our backdoor attack designed on the basis of visual prompting. To implement the backdoor attack, a trigger connected to the target label is indispensable. We first introduce the trigger which is an additive noise initialize from a uniform distribution, denoted as $\delta \in R^{w\times h}$, in which $w$ and $h$ are the width and height of the input image $x$, respectively. So a triggered image can be formalized as $x+\delta$. To achieve the goal of human imperceptibility, we also introduce $l^{\infty}$ restriction on $\delta$ to keep the trigger stealthy. Accordingly, the maximum of pixel change is less than $\epsilon$.

Next, in order to implement the switchable mode, we additionally introduce an extra token as the switch on the basis of the original visual prompts. The switch token is used for toggling the backdoor mode on or off. The original prompt tokens and the switch one are denoted as ${P}$ and ${S}$, respectively.

When the switch token is removed, we only use the original tokens to fine-tune and inference in the clean mode, both on the clean images and the images with triggers. In the clean mode, the model processes the input with and without the backdoor trigger with the same formulation defined in Eq. (\ref{eq:1}).
As a contrast, when the switch token is attached, the backdoor mode is turned on. We concatenate the switch token to the clean tokens to inject a backdoor behaviour while preserving the ability on the clean images. We formulate this process for the layer $L_1$ as follow:
\begin{equation}
\begin{aligned}
\relax [c_{1},{Z_{1}},{Z'_{1}},{E_{1}}]=L_{1}(c_{0},{P},S,E_{0}), 
\label{eq:switch}
\end{aligned}
\end{equation}
where $E_{0}=\text{Patch\_Emb}(x\!)$ for clean input or $E_{0}=\text{Patch\_Emb}(x\!+\!\delta)$ for backdoor input. $Z'_{1}$ is the features computed by the switch token. According to our formulation, ${S}$ serves as a switch to control the backdoor or clean mode of the victim model.

\subsection{Objective Functions}
As mentioned above, different goals are needed to be accomplished in two modes, which is the key part of SWARM. Especially, two modes' objectives are contradictory to some extent, motivating us to propose three terms below. \cref{fig:method} summarizes the learning objective in our SWARM.

\noindent  \textbf{Clean loss.} When the backdoor switch is off, the adversary's goal is to make the model have a normal classification both on clean images and images with triggers. The parameters of the clean tokens $P$ and the trigger $\delta$ are trained in this process by minimizing the empirical classification loss:
\begin{equation}
\begin{aligned}
\mathcal{L}_{cle}\left ( P,\delta \right )= \mathbb{E}_{(x,y) \sim \mathcal{D}}& [\ell(P,x,y)+\ell(P,x+\delta,y)]\\
s.t. &\left \| \delta \right \| _{\infty } \le \epsilon,
\end{aligned}
\end{equation}
where $\ell(\cdot)$ calculates the cross-entropy loss.
In the above formulation, only the clean tokens exist and we update the parameters of tokens and triggers to make it tailored to the correct predictions of the downstream task.

\noindent  \textbf{Backdoor loss.} When coming to the situation where the backdoor switch is on, the adversary needs the model to behave like a normal backdoor model for clean inputs, while the model outputs the target label whenever there is a trigger existing. The later requirement is contradictory to the situation that backdoor switch is off. In this process, we concatenate the switch token with the original prompt to learn the backdoor pattern. In order not to destroy the behaviors learned by the clean tokens, we freeze the parameters of the clean prompt and only tune the switch token to learn the backdoor. We set the target label as $t$ and this process can also be trained by minimizing the empirical classification loss which can be formulated as:
\begin{equation}
\begin{aligned}
\mathcal{L}_{bd}\left (S,\delta \right )\!=\! \mathbb{E}_{(x,y) \sim \mathcal{D}}&[\ell(P, S,x,y)\!+\!\ell(P,S,x\!+\!\delta,t)]\\
s.t. &\left \| \delta \right \| _{\infty } \le \epsilon,
\end{aligned}
\end{equation}
In the equation, the switch token $S$ is updated in this process while we keep the parameters of the original prompt $P$ frozen. 
Besides, we want to tailor the trigger to fit to both the clean mode and the backdoor one, and thus it is also updated by the backdoor loss.

\noindent  \textbf{Cross-mode feature distillation loss.} 
The clean and backdoor loss terms contribute to two separate goals in clean and backdoor modes, respectively. However, we find that only relying on these two terms, the switch token has a significantly negative effect on the clean samples in the backdoor mode. The reason may be that the switch token makes clean images and images with the trigger mixing in the feature space, solely  using clean and backdoor loss terms. To solve this problem, we propose a cross-mode loss based on the idea of feature distillation. It can be formulated as follows:
\begin{equation}
\begin{aligned}
\mathcal{L}_{cs}\left (S \right )&= \mathbb{E}_{(x,y) \sim \mathcal{D}} ||F_f(P,x)-F_f(P,S,x)||_2,
\end{aligned}
\end{equation}
where $F_f(\cdot)$ outputs the feature before the last classifier of the input sample $x$. Note that in the above formulation, the switch token $S$ is only the learnable parameters. The key idea of $\mathcal{L}_{cs}$ is to minimize the distance between features of inputs with the switch token and these of inputs without the switch token.  
Accordingly, with a trade-off parameter $\lambda$, the overall objective of our SWARM can be formulated as:
\begin{equation}
\begin{aligned}
\mathcal{L}_{total}=\mathcal{L}_{cle}+\mathcal{L}_{bd}+\lambda \mathcal{L}_{cs}.
\end{aligned}
\end{equation}

\subsection{Learning Strategy}
We adopt a learning strategy to realize two different modes in our SWARM. In one iteration step, we first use the clean loss to update the clean tokens and trigger. And then, we freeze the clean tokens and add the switch token to the input. We use backdoor loss and cross-mode feature distillation loss to update the switch token and trigger. Therefore, we need twice forward and backward propagations in one step to optimize the parameters.
\section{Experiments}

\begin{table*}[]
\centering
\caption{The main results (\%) on VTAB-1k \cite{zhai2019large} dataset collection. SWARM is competitive with four advanced backdoor attacks in terms of BA, and meanwhile reaches high ASRs which exceeds 95\%. We have marked the best BA and ASR on 5 backdoor attacks with bold scores while underlined scores are the second-best performance.}
\label{table:main}
\scalebox{0.8}{
\begin{tabular}{c|cccccccc|c|cccc}
\toprule
                Attack$\rightarrow$      & \multicolumn{1}{c|}{No Attack} & \multicolumn{2}{c|}{BadNets} & \multicolumn{2}{c|}{Blended} & \multicolumn{2}{c|}{WaNet} & \multicolumn{2}{c|}{ISSBA}  & \multicolumn{2}{c||}{SWARM-B} & \multicolumn{2}{c}{SWARM-C}    \\ \hline
Datasets-VTAB$\downarrow$, Metric$\rightarrow$   & \multicolumn{1}{c|}{BA}  & \multicolumn{1}{c|}{BA}           & \multicolumn{1}{c|}{ASR}          & \multicolumn{1}{c|}{BA}           & \multicolumn{1}{c|}{ASR}          & \multicolumn{1}{c|}{BA}          & \multicolumn{1}{c|}{ASR}         & \multicolumn{1}{c|}{BA}          & \multicolumn{1}{c|}{ASR}    & \multicolumn{1}{c|}{BA}    & \multicolumn{1}{c||}{ASR}  & \multicolumn{1}{c|}{BA}    & \multicolumn{1}{c}{BA-T} \\ \hline
CIFAR-100     & \multicolumn{1}{c|}{77.27}       & \multicolumn{1}{c|}{67.57}        & \multicolumn{1}{c|}{86.07}        & \multicolumn{1}{c|}{64.82}        & \multicolumn{1}{c|}{85.65}        & \multicolumn{1}{c|}{65.72 }      & \multicolumn{1}{c|}{83.72}       & \multicolumn{1}{c|}{\underline{72.87}}      & \multicolumn{1}{c|}{\textbf{99.28}} & \multicolumn{1}{c|}{\textbf{76.36}} & \multicolumn{1}{c||}{\underline{96.96}} & \multicolumn{1}{c|}{76.41} & \multicolumn{1}{c}{76.38} \\ \hline
Caltech101     & \multicolumn{1}{c|}{83.89}      & \multicolumn{1}{c|}{46.11}        & \multicolumn{1}{c|}{50.93}        & \multicolumn{1}{c|}{41.51}        & \multicolumn{1}{c|}{54.77}        & \multicolumn{1}{c|}{52.98}       & \multicolumn{1}{c|}{48.33}       & \multicolumn{1}{c|}{\underline{79.85}}       & \multicolumn{1}{c|}{\underline{89.99}}  & \multicolumn{1}{c|}{\textbf{82.63}} & \multicolumn{1}{c||}{\textbf{96.58}} & \multicolumn{1}{c|}{84.32} & \multicolumn{1}{c}{84.01}\\ \hline
DTD       & \multicolumn{1}{c|}{65.90}          & \multicolumn{1}{c|}{37.23}        & \multicolumn{1}{c|}{73.94}        & \multicolumn{1}{c|}{34.10}        & \multicolumn{1}{c|}{60.85}        & \multicolumn{1}{c|}{35.32}       & \multicolumn{1}{c|}{62.29}       & \multicolumn{1}{c|}{20.53}       & \multicolumn{1}{c|}{87.82}  & \multicolumn{1}{c|}{\textbf{62.11}} & \multicolumn{1}{c||}{\textbf{95.11}} & \multicolumn{1}{c|}{63.67} & \multicolumn{1}{c}{63.99}\\ \hline
Flowers102      & \multicolumn{1}{c|}{97.48}     & \multicolumn{1}{c|}{\textbf{94.73}}        & \multicolumn{1}{c|}{\underline{91.15}}        & \multicolumn{1}{c|}{91.61}        & \multicolumn{1}{c|}{80.01}        & \multicolumn{1}{c|}{80.40}       & \multicolumn{1}{c|}{28.17}       & \multicolumn{1}{c|}{84.23}       & \multicolumn{1}{c|}{88.55}    & \multicolumn{1}{c|}{\underline{93.53}} & \multicolumn{1}{c||}{\textbf{96.99}} & \multicolumn{1}{c|}{96.80} & \multicolumn{1}{c}{96.93} \\ \hline
Pets       & \multicolumn{1}{c|}{87.52}     & \multicolumn{1}{c|}{\underline{81.49}}        & \multicolumn{1}{c|}{\underline{87.52}}       & \multicolumn{1}{c|}{81.90}        & \multicolumn{1}{c|}{79.56}        & \multicolumn{1}{c|}{73.86}       & \multicolumn{1}{c|}{34.94 }      & \multicolumn{1}{c|}{73.67}       & \multicolumn{1}{c|}{87.46}    & \multicolumn{1}{c|}{\textbf{86.02}} & \multicolumn{1}{c||}{\textbf{98.53}} & \multicolumn{1}{c|}{86.64} & \multicolumn{1}{c}{86.43}\\ \hline
SVHN    & \multicolumn{1}{c|}{68.76}    & \multicolumn{1}{c|}{61.39 }       & \multicolumn{1}{c|}{90.04 }       & \multicolumn{1}{c|}{62.83}        & \multicolumn{1}{c|}{91.79}        & \multicolumn{1}{c|}{50.58}       & \multicolumn{1}{c|}{33.09}       & \multicolumn{1}{c|}{\underline{66.63}}       & \multicolumn{1}{c|}{\textbf{99.24}}     & \multicolumn{1}{c|}{\textbf{67.72}} & \multicolumn{1}{c||}{\underline{96.05}}        & \multicolumn{1}{c|}{67.84} & \multicolumn{1}{c}{68.81}\\ \hline
Sun397     & \multicolumn{1}{c|}{47.83 }          & \multicolumn{1}{c|}{29.35}        & \multicolumn{1}{c|}{73.92}        & \multicolumn{1}{c|}{26.02}        & \multicolumn{1}{c|}{57.03}        & \multicolumn{1}{c|}{24.92}       & \multicolumn{1}{c|}{71.14}       & \multicolumn{1}{c|}{\underline{35.76}}       & \multicolumn{1}{c|}{\underline{92.81}}     & \multicolumn{1}{c|}{\textbf{43.53}} & \multicolumn{1}{c||}{\textbf{96.53}}  & \multicolumn{1}{c|}{47.41} & \multicolumn{1}{c}{45.40} \\ \hline
Patch Camelyon   & \multicolumn{1}{c|}{75.01}   & \multicolumn{1}{c|}{69.62}        & \multicolumn{1}{c|}{70.63}        & \multicolumn{1}{c|}{67.15}        & \multicolumn{1}{c|}{75.73}        & \multicolumn{1}{c|}{63.62 }      & \multicolumn{1}{c|}{82.71}       & \multicolumn{1}{c|}{\underline{72.98}}       & \multicolumn{1}{c|}{\underline{96.43}}     & \multicolumn{1}{c|}{\textbf{76.65}} & \multicolumn{1}{c||}{\textbf{96.56}} & \multicolumn{1}{c|}{78.37} & \multicolumn{1}{c}{77.83} \\ \hline
EuroSAT    & \multicolumn{1}{c|}{92.96}          & \multicolumn{1}{c|}{90.74}        & \multicolumn{1}{c|}{\underline{98.96}}        & \multicolumn{1}{c|}{90.37}        & \multicolumn{1}{c|}{95.89}        & \multicolumn{1}{c|}{77.17}       & \multicolumn{1}{c|}{27.72}       & \multicolumn{1}{c|}{\underline{91.24}}       & \multicolumn{1}{c|}{\textbf{99.67}}    & \multicolumn{1}{c|}{\textbf{91.94}} & \multicolumn{1}{c||}{96.52}    & \multicolumn{1}{c|}{92.09} & \multicolumn{1}{c}{91.30} \\ \hline
Clevr/count     & \multicolumn{1}{c|}{45.73}     & \multicolumn{1}{c|}{42.36}        & \multicolumn{1}{c|}{\textbf{100.00}}       & \multicolumn{1}{c|}{42.77}        & \multicolumn{1}{c|}{\textbf{100.00}}       & \multicolumn{1}{c|}{38.67}       & \multicolumn{1}{c|}{96.19}       & \multicolumn{1}{c|}{\underline{43.70}}       & \multicolumn{1}{c|}{\textbf{100.00}}    & \multicolumn{1}{c|}{\textbf{44.83}} & \multicolumn{1}{c||}{\underline{99.98}}   & \multicolumn{1}{c|}{45.60} & \multicolumn{1}{c}{45.53} \\ \hline
Clevr/distance   & \multicolumn{1}{c|}{54.13}    & \multicolumn{1}{c|}{\textbf{53.89}}        & \multicolumn{1}{c|}{99.98}        & \multicolumn{1}{c|}{51.39}        & \multicolumn{1}{c|}{\textbf{100.00}}       & \multicolumn{1}{c|}{40.75}       & \multicolumn{1}{c|}{64.23 }      & \multicolumn{1}{c|}{\underline{52.26}}       & \multicolumn{1}{c|}{\textbf{100.00}}     & \multicolumn{1}{c|}{49.37} & \multicolumn{1}{c||}{\underline{99.99}}     & \multicolumn{1}{c|}{50.98} & \multicolumn{1}{c}{50.37} \\ \hline
DMLab       & \multicolumn{1}{c|}{36.92}         & \multicolumn{1}{c|}{34.04}        & \multicolumn{1}{c|}{\underline{99.51}}       & \multicolumn{1}{c|}{34.41}        & \multicolumn{1}{c|}{99.48}        & \multicolumn{1}{c|}{33.87}       & \multicolumn{1}{c|}{75.70}       & \multicolumn{1}{c|}{\underline{34.18}}       & \multicolumn{1}{c|}{\textbf{99.56}}    & \multicolumn{1}{c|}{\textbf{34.34}} & \multicolumn{1}{c||}{97.39}   & \multicolumn{1}{c|}{34.97} & \multicolumn{1}{c}{34.77} \\ \hline
KITTI      & \multicolumn{1}{c|}{66.38}          & \multicolumn{1}{c|}{60.90}        & \multicolumn{1}{c|}{\textbf{99.72}}        & \multicolumn{1}{c|}{62.59 }       & \multicolumn{1}{c|}{96.06 }       & \multicolumn{1}{c|}{63.71}       & \multicolumn{1}{c|}{92.12}       & \multicolumn{1}{c|}{\underline{64.70}}      & \multicolumn{1}{c|}{96.77}    & \multicolumn{1}{c|}{\textbf{65.96}} & \multicolumn{1}{c||}{\underline{98.87}}    & \multicolumn{1}{c|}{69.20} & \multicolumn{1}{c}{62.59}  \\ \hline
dSprites/location  & \multicolumn{1}{c|}{70.78}  & \multicolumn{1}{c|}{62.23}        & \multicolumn{1}{c|}{\textbf{100.00}}       & \multicolumn{1}{c|}{63.80 }       & \multicolumn{1}{c|}{\underline{99.96}}        & \multicolumn{1}{c|}{53.12}       & \multicolumn{1}{c|}{24.92}       & \multicolumn{1}{c|}{\underline{68.57}}       & \multicolumn{1}{c|}{99.84}    & \multicolumn{1}{c|}{\textbf{68.83}} & \multicolumn{1}{c||}{99.79}    & \multicolumn{1}{c|}{69.97} & \multicolumn{1}{c}{69.29}\\ \hline
dSprites/orientation & \multicolumn{1}{c|}{35.39}  & \multicolumn{1}{c|}{26.27 }       & \multicolumn{1}{c|}{\textbf{99.94}}        & \multicolumn{1}{c|}{29.55}        & \multicolumn{1}{c|}{\underline{99.87}}        & \multicolumn{1}{c|}{24.91}       & \multicolumn{1}{c|}{48.62}       & \multicolumn{1}{c|}{\underline{33.82}}       & \multicolumn{1}{c|}{99.83}     & \multicolumn{1}{c|}{\textbf{36.58}} & \multicolumn{1}{c||}{99.62}   & \multicolumn{1}{c|}{36.39} & \multicolumn{1}{c}{36.41} \\ \hline
SmallNORB/azimuth  & \multicolumn{1}{c|}{11.96}  & \multicolumn{1}{c|}{9.31}         & \multicolumn{1}{c|}{96.40}        & \multicolumn{1}{c|}{7.65}         & \multicolumn{1}{c|}{79.25}        & \multicolumn{1}{c|}{7.72}        & \multicolumn{1}{c|}{77.23}       & \multicolumn{1}{c|}{\textbf{13.42}}       & \multicolumn{1}{c|}{\textbf{100.00}}   & \multicolumn{1}{c|}{\underline{9.95}}  & \multicolumn{1}{c||}{\underline{99.06}}       & \multicolumn{1}{c|}{13.55} & \multicolumn{1}{c}{13.43} \\ \hline
SmallNORB/elevation  & \multicolumn{1}{c|}{27.29}  & \multicolumn{1}{c|}{26.16}        & \multicolumn{1}{c|}{86.36 }       & \multicolumn{1}{c|}{27.85}        & \multicolumn{1}{c|}{85.08}        & \multicolumn{1}{c|}{22.05}       & \multicolumn{1}{c|}{47.41}       & \multicolumn{1}{c|}{\underline{30.20}}       & \multicolumn{1}{c|}{\textbf{99.89}}    & \multicolumn{1}{c|}{\textbf{30.77}} & \multicolumn{1}{c||}{\underline{99.79}}    & \multicolumn{1}{c|}{31.36} & \multicolumn{1}{c}{30.49} \\ \midrule

Average    & \multicolumn{1}{c|}{61.48}    & \multicolumn{1}{c|}{52.55}         & \multicolumn{1}{c|}{88.53}  & \multicolumn{1}{c|}{51.78}   & \multicolumn{1}{c|}{84.76}  &  \multicolumn{1}{c|}{47.61} &  \multicolumn{1}{c|}{58.74}   &  \multicolumn{1}{c|}{\underline{55.21}}  &  \multicolumn{1}{c|}{\underline{96.32}} &  \multicolumn{1}{c|}{\textbf{59.95}}  &  \multicolumn{1}{c||}{\textbf{97.90}}   &  \multicolumn{1}{c|}{61.50}  &  \multicolumn{1}{c}{60.82}\\

\bottomrule
\end{tabular}
}
\begin{tablenotes}
\footnotesize
\item[1]SWARM-B: The switch token is added and the model is under backdoor mode.
\item[2]SWARM-C: The switch token is removed and the model is under clean mode.Therefore, the images with triggers are still have normal performance.
\end{tablenotes}
\vspace{-1em}
\end{table*}

\begin{figure}[]
    \centering
    \includegraphics[width=0.4\textwidth]{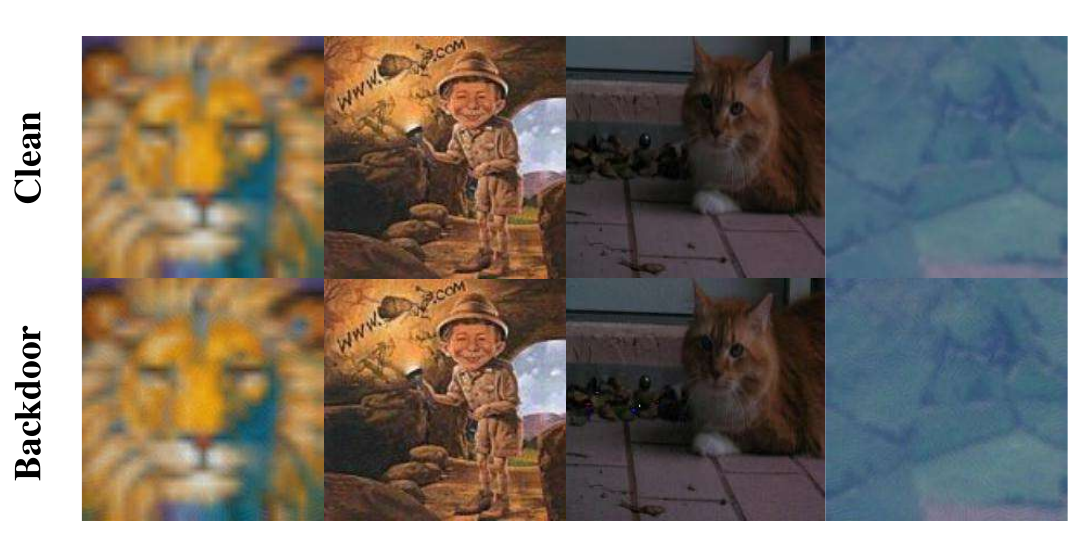}
    \caption{Visualization of clean and backdoor images.}
    \vspace{-1.5em}
\end{figure}

In this section, we will evaluate the performance of SWARM across various vision datasets, the impact of different hyper-parameters, and its robustness to backdoor detection and mitigation.

\subsection{Experimental Setup}
\textbf{Datasets and models.}
We evaluate SWARM on datasets from the VTAB-1k \cite{zhai2019large} benchmark. Concretely, VTAB-1k is a collection of diverse visual classification tasks, which can be divided into three groups: Natural tasks contain natural images captured by standard cameras; Specialized tasks contain images captured by special equipment such as medical and satellite imagery; and Structured tasks require the models to have the geometric comprehension. In this collection, each task contains 1000 training samples and we use the provided split of the train set to evaluate our SWARM. In this scenario, only 800 samples are used for the training while the remaining 200 samples are used for validation. Besides, we select Vision Transformer (ViT) \cite{dosovitskiy2020image} which is pre-trained on Imagenet-21K \cite{deng2009imagenet} as the main target model.

\noindent \textbf{Baselines and attack settings.}
We choose 4 existing backdoor attacks as our baselines: BadNets \cite{gu2017badnets}, Blended \cite{chen2017targeted}, WaNet \cite{nguyen2021wanet} and ISSBA \cite{li2021invisible}. 
We adapt these attacks by setting the prompts as the only learnable parameters.
We maintain the default settings following their papers to ensure the performance. 
We set the clean tokens to 50 in all cases and other settings of the visual prompt learning are drawn from \cite{jia2022visual}. 
For human imperceptibility, the $\epsilon$ is set to 4. 
The hyper-parameter $\lambda$ is set to 100 as the default.

\noindent \textbf{Evaluation metrics.} There are two modes that need to be evaluated, where we use SWARM-C and SWARM-B to denote the model in the clean and backdoor mode, respectively. In clean mode, we should evaluate the model's performance under the clean images and images with triggers. Therefore, we use Benign Accuracy (BA) and Benign Accuracy with Triggers (BA-T) to measure the performance in the clean mode. While in the backdoor mode, we follow previous backdoor studies \cite{nguyen2021wanet,li2022backdoor,yang2023backdoor,gao2023backdoor}, which use Benign Accuracy (BA) and Attack Success Rate (ASR) to measure the backdoor attack. Specifically, higher values of these metrics indicate the better performance.

\subsection{Main Results}
In this section, we perform our SWARM on VTAB-1k and present the results in \cref{table:main}.

\noindent  \textbf{SWARM-C correctly classifies clean images and triggered images.} As observed in \cref{table:main}, SWARM-C can achieve comparable performance on both clean images and triggered images compared to the no-attack situation among all the datasets. In most cases, SWARM performs a minor accuracy drop of less than $2\%$. Meanwhile, no performance decline exists between triggered and clean images, indicating that even though the input images are triggered, it is difficult for the victims to detect performance differences under this mode. In some cases, SWARM-C outperforms the no-attack situation, ensuring its competitiveness.

\noindent  \textbf{SWARM-B correctly classifies clean images.} After the switch token is added, the model is changed to the backdoor mode. In this situation, SWARM acts as a normal backdoor model which will manipulate the prediction result whenever there are backdoor triggers. As is shown in \cref{table:main}, SWARM-B achieves the best benign accuracy among all backdoor attacks, which has less than a $2\%$ drop compared to the no-attack situation in most cases. The average of the SWARM's BA is also the highest in these methods.

\noindent  \textbf{SWARM-B achieves high attack success rates.} We can see from \cref{table:main} that SWARM shows promising performance in terms of ASR. Specifically, SWARM achieves high ASRs ($>95\%$) on all datasets and a $97.90\%$ on average. Moreover, $97.90\%$ is the highest average ASR value among all backdoor attacks. Although ISSBA can achieve a very competitive ASR in some cases, it has some bad performance on specific datasets with less than 90\% ASR.


\subsection{Ablation Study}
\textbf{SWARM on different backbones.}
\begin{table}[]
\caption{Results of SWARM on different backbones. It has the same performance comparing to the ViT.}
\scalebox{0.8}{
\begin{tabular}{c|c|cccc}
\toprule
Attack$\rightarrow$                  & \multicolumn{1}{c|}{No Attack} & \multicolumn{2}{c||}{SWARM-B}    & \multicolumn{2}{c}{SWARM-C} \\ \hline
Backbones$\downarrow$, Metric$\rightarrow$  & \multicolumn{1}{c|}{BA}      & \multicolumn{1}{c|}{BA}    & \multicolumn{1}{c||}{ASR}  & \multicolumn{1}{c|}{BA}    & \multicolumn{1}{c}{BA-T}  \\ \hline
ViT               & \multicolumn{1}{c|}{77.27}   & \multicolumn{1}{c|}{76.36} & \multicolumn{1}{c||}{96.96}  & \multicolumn{1}{c|}{76.41} & \multicolumn{1}{c}{76.38}\\ \hline
Swin-B            & \multicolumn{1}{c|}{72.62}   & \multicolumn{1}{c|}{70.11} & \multicolumn{1}{c||}{97.66}  & \multicolumn{1}{c|}{71.11} & \multicolumn{1}{c}{70.72} \\ \hline
ConvNeXt-Base     & \multicolumn{1}{c|}{73.31}   & \multicolumn{1}{c|}{73.43} & \multicolumn{1}{c||}{96.64}  & \multicolumn{1}{c|}{75.51} & \multicolumn{1}{c}{76.24} \\ \midrule
\end{tabular}
}
\end{table}
In this part, we evaluate our methods on different backbones. Visual prompts can not only be added to ViT but also they can be used for other backbones. As shown in Table 2, we further experiment with our SWARM on Swin Transformer \cite{liu2021swin} and ConvNeXt \cite{liu2022convnet}. These pre-trained models are all pre-trained on the Imagenet-21K \cite{deng2009imagenet} and then adapted with our SWARM. Our method still has $96\%+$ ASRs for these two backbones. It demonstrates SWARM's effectiveness with no regard to the upstream backbones. 

\noindent \textbf{Effect of the number of switch tokens.}
\begin{figure}[]
    \centering
    \includegraphics[width=0.3\textwidth]{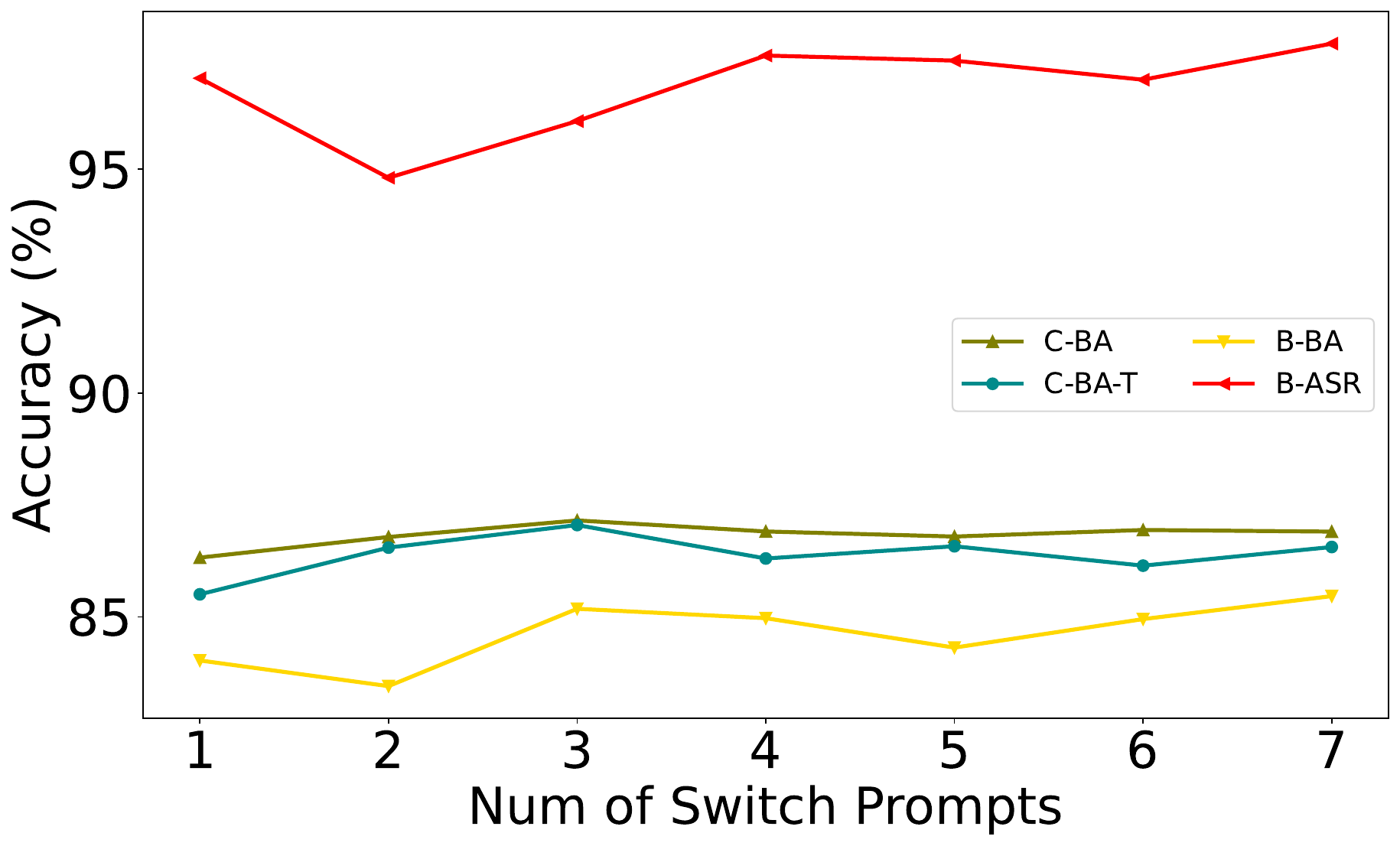}
    \caption{The effect of increasing the numbers of switch tokens.}
    \label{fig:numofswitch}
    \vspace{-1.5em}
\end{figure}
In our SWARM, we adopt a single token as the switch. Here, we investigate the impact of varying the number of switch tokens on three different datasets: CIFAR100 \cite{krizhevsky2009learning}, Flowers \cite{nilsback2008automated}, and Pets \cite{parkhi2012cats}. As demonstrated in \cref{fig:numofswitch}, performance does not improve with an increasing number of switch tokens, suggesting that one switch token suffices for our method.


\noindent \textbf{Effect of $\lambda$.}
\begin{figure}[]
    \centering
    \includegraphics[width=0.3\textwidth]{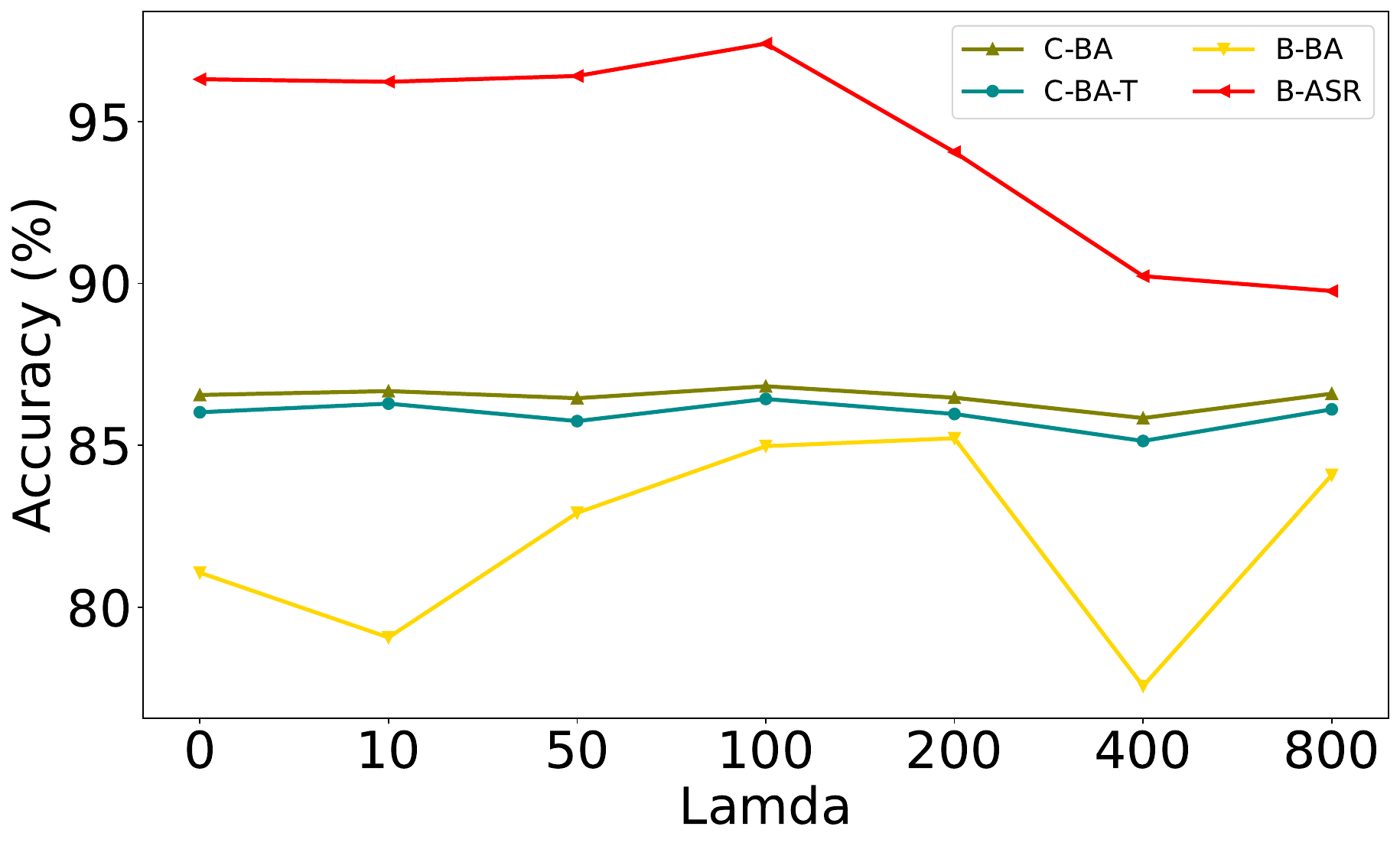}
        \caption{The effect of increasing the $\lambda$.}
    \label{fig:lambda}
    \vspace{-1em}
\end{figure}
In our SWARM, the weight of cross-mode feature distillation loss $\lambda$ is set to 100. We further explore the impact of $\lambda$ on the three previously mentioned datasets. As depicted in \cref{fig:lambda}, performance remains robust when $\lambda$ does not exceed 200, with 100 being the most suitable trade-off parameter.

\noindent \textbf{Effect of the switch token and the cross-mode distillation loss.}
\begin{table}[]
\centering
\caption{Effect of the switch token $S$ and the cross-mode distillation loss $\mathcal{L}_{cs}$  on three datasets.}
\scalebox{0.85}{
\begin{tabular}{c|c|cc|cc}
\toprule
Datasets                  & Mode              & \multicolumn{2}{c|}{SWARM-B}       & \multicolumn{2}{c}{SWARM-C}        \\ \hline
/                         & Metric            & \multicolumn{1}{c|}{BA}    & ASR   & \multicolumn{1}{c|}{BA}    & BA-T \\ \hline
 & w/o $S$ & \multicolumn{1}{c|}{36.64} & 66.38 & \multicolumn{1}{c|}{36.64} & 25.91 \\ \cline{2-6} 
    {CIFAR100}                      & w/o $\mathcal{L}_{cs}$  & \multicolumn{1}{c|}{69.75} & 98.09 & \multicolumn{1}{c|}{76.03} & 74.91 \\ \cline{2-6} 
                          & w/ all            & \multicolumn{1}{c|}{76.36} & 96.96 & \multicolumn{1}{c|}{76.41} & 76.38 \\ \hline
  & w/o $S$ & \multicolumn{1}{c|}{80.18} & 70.28 & \multicolumn{1}{c|}{80.18} & 23.52 \\ \cline{2-6} 
    {Flowers}                      & w/o $\mathcal{L}_{cs}$  & \multicolumn{1}{c|}{91.09} & 95.33 & \multicolumn{1}{c|}{91.09} & 95.33 \\ \cline{2-6} 
                          & w/ all            & \multicolumn{1}{c|}{93.53} & 96.99 & \multicolumn{1}{c|}{96.80}  & 96.93 \\ \hline
     & w/o $S$ & \multicolumn{1}{c|}{76.45} & 68.68 & \multicolumn{1}{c|}{76.45} & 25.92 \\ \cline{2-6} 
     {Pets}                     & w/o $\mathcal{L}_{cs}$  & \multicolumn{1}{c|}{82.37} & 95.50  & \multicolumn{1}{c|}{87.19} & 86.92 \\ \cline{2-6} 
                          & w/ all            & \multicolumn{1}{c|}{86.02} & 98.53 & \multicolumn{1}{c|}{86.64} & 86.43 \\ \bottomrule
\end{tabular}
}
\label{table:ablation}
\vspace{-1em}
\end{table}
As illustrated in \cref{table:ablation}, we demonstrate the indispensability of each component. Without the switch token, SWARM exhibits poor performance across both modes. In the absence of cross-mode feature distillation loss, SWARM behaves normally in clean mode. However, the model cannot accurately classify clean images, resulting in a 10\% drop in terms of BA in backdoor mode. Besides, we perform the ablation on trigger learning in Appendix. 


\begin{figure}[]
  \centering            
  \subfloat[SWARM-C-Clean Images]
  {
      \label{fig:subfig1_1}\includegraphics[width=0.23\textwidth]{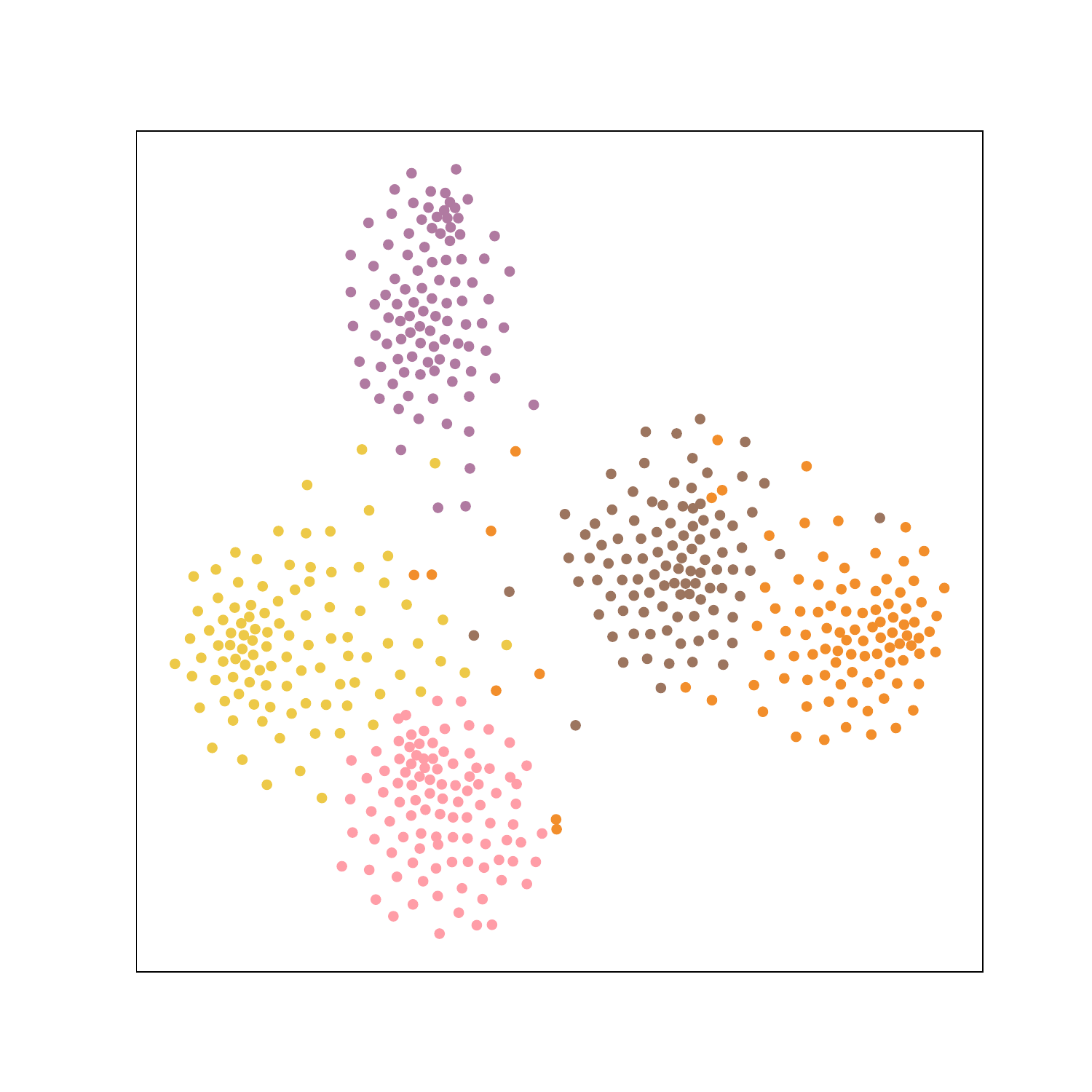}
  }
  \subfloat[SWARM-C-Triggered Images]
  {
      \label{fig:subfig2_1}\includegraphics[width=0.23\textwidth]{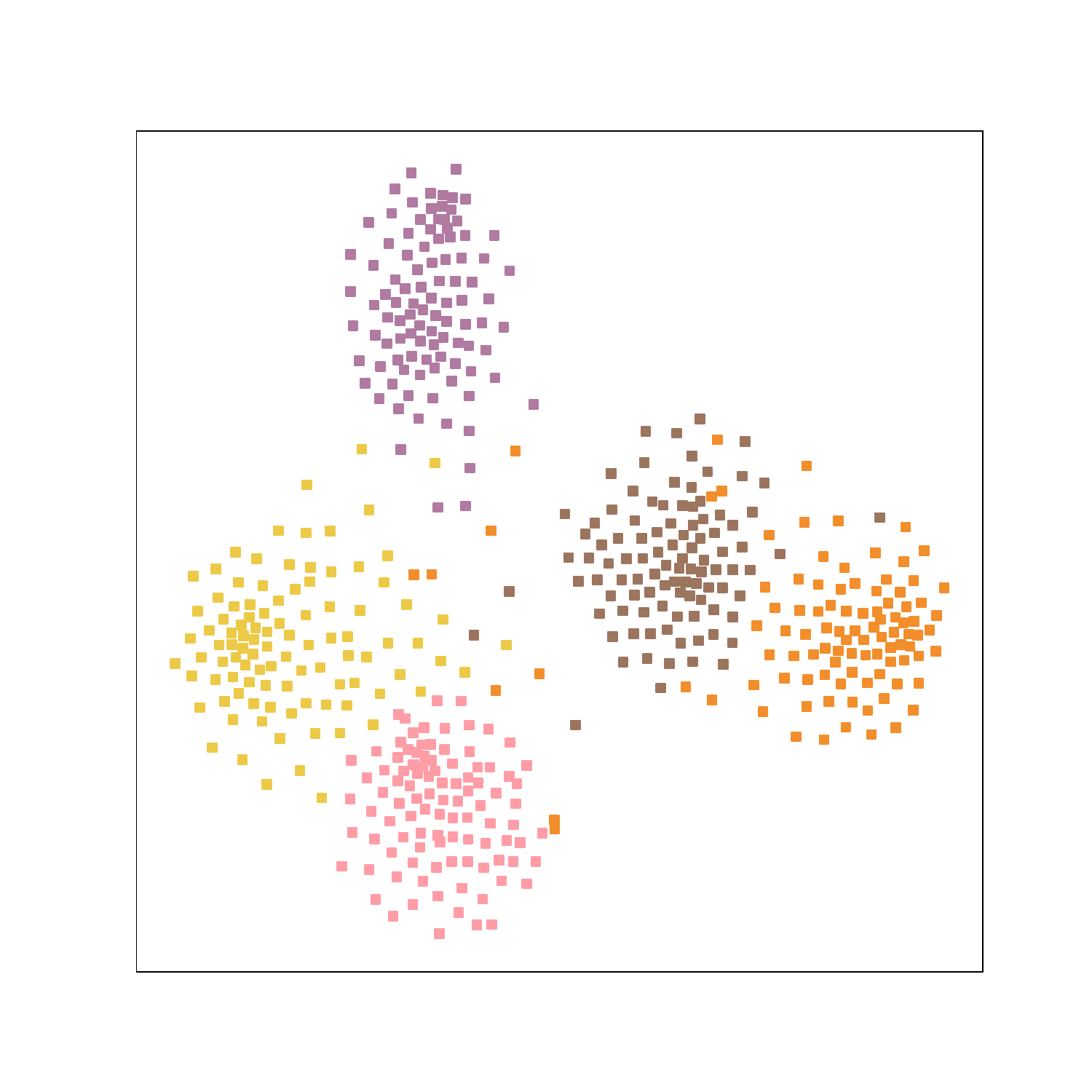}
  }
  \vfill
  \subfloat[SWARM-B-Clean Images]
  {
      \label{fig:subfig1_2}\includegraphics[width=0.23\textwidth]{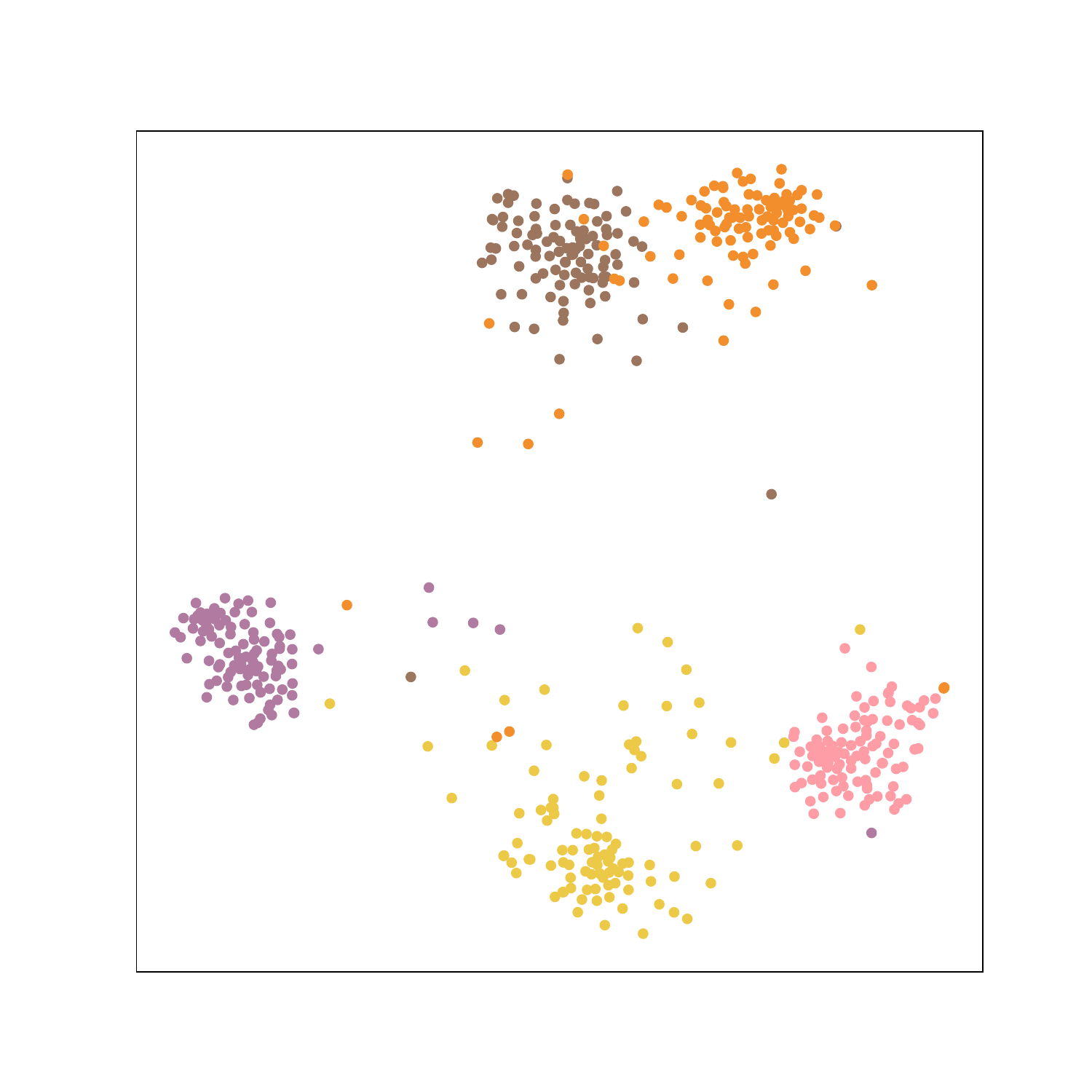}
  }
  \subfloat[SWARM-B-Triggered Images]
  {
      \label{fig:subfig2_2}\includegraphics[width=0.23\textwidth]{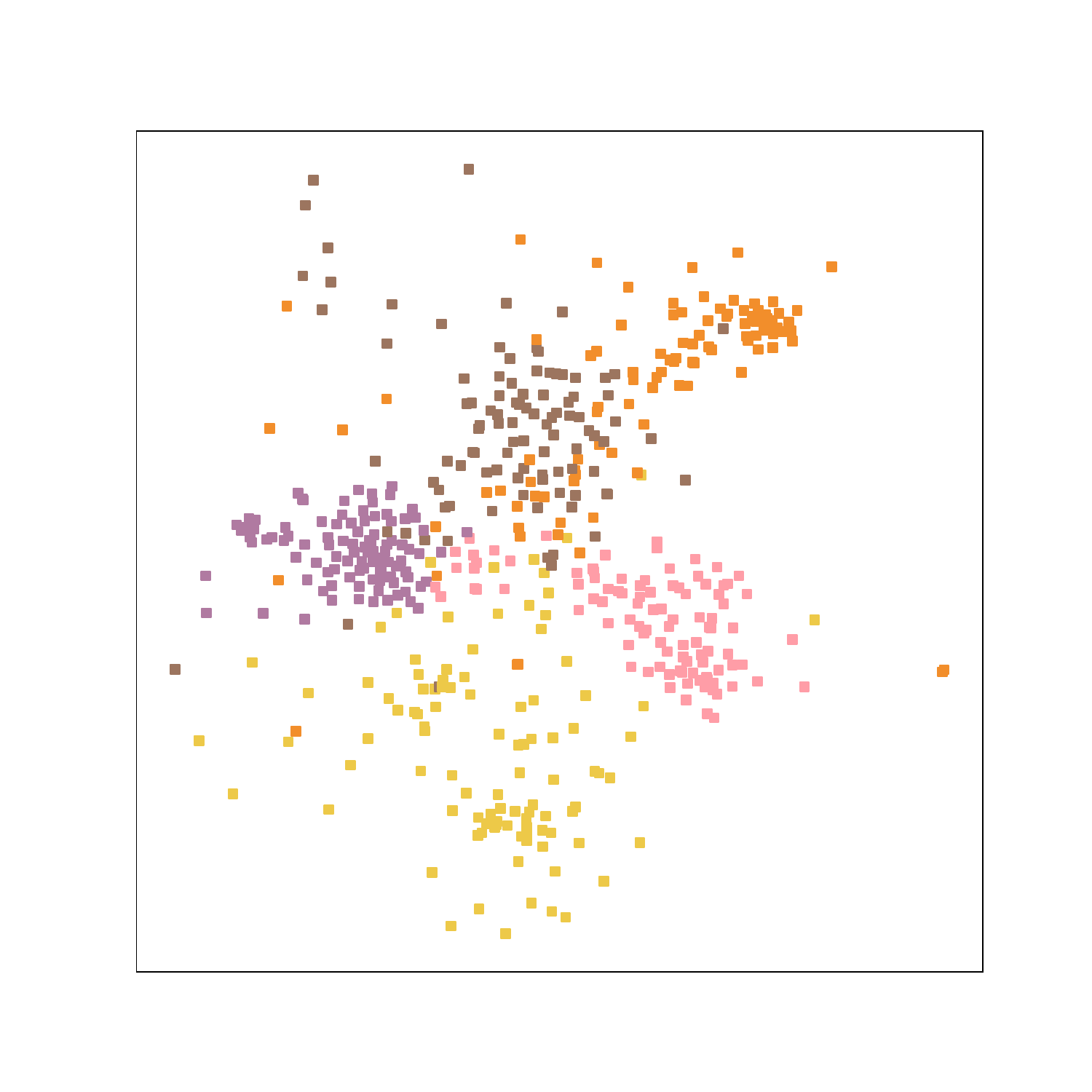}
  }
  \caption{The t-SNE visualization of features extracted by SWARM. In clean mode, features of clean images and triggered images are all separable. In backdoor mode, features of clean images are separable while these of triggered images cluster together.}
  \label{fig:visualization}
  \vspace{-1em}
\end{figure}

\subsection{Visualization}
As shown in \cref{fig:visualization}, we visualize the features obtained from the SWARM model. In the clean mode, we can observe that the clean features and the triggered have almost the same pattern and they are all separable, which explains the clean performance on the triggered images. In the backdoor mode, clean images' features are still separable which indicates the good prediction results on benign accuracy. In contrast, for triggered images' features in the backdoor mode, the situation is poles apart, i.e., the borders of the features are not as clear as the clean ones. The triggered images gather together so the classifier naturally makes the target predictions on these inputs. This phenomenon demonstrates our method's rationality and it is a worthy point to be further researched that only a small portion of parameters in the switch token can achieve such an obvious change in behaviors during the adaption.


\subsection{Robustness to Backdoor Detection}
In this section, we choose three backdoor detection methods\cite{gao2023backdoor} to check the attacks' stealthiness. They are Scale-Up \cite{guo2023scale}, TeCo \cite{liu2023detecting} and STRIP \cite{gao2019strip} (the results of STRIP are shown in Appendix). Following  existing detection-based backdoor defenses \cite{gao2021design,guo2021aeva,huang2019neuroninspect}, we use Area Under Receiver Operating Curve (AUROC) \cite{fawcett2006introduction} as the metric, which is widely used to measure the trade-off between the false positive rate for clean samples and true positive rate for triggered samples. 
Besides, we adopt another metric called ASR-D. To calculate it, we first select out the clean samples categorized by the detection method and then use these samples to calculate the Attack Success Rate on the backdoored model. The higher ASR-D, the better the stealthiness of the backdoor attack since it means that detection methods can not detect the triggered samples. According to our threat model, we set the SWARM to clean mode when we meet the detection method and we test the attack performance of our method in the backdoor mode.

\begin{figure}[!t]
    \centering
    \includegraphics[width=0.47\textwidth]{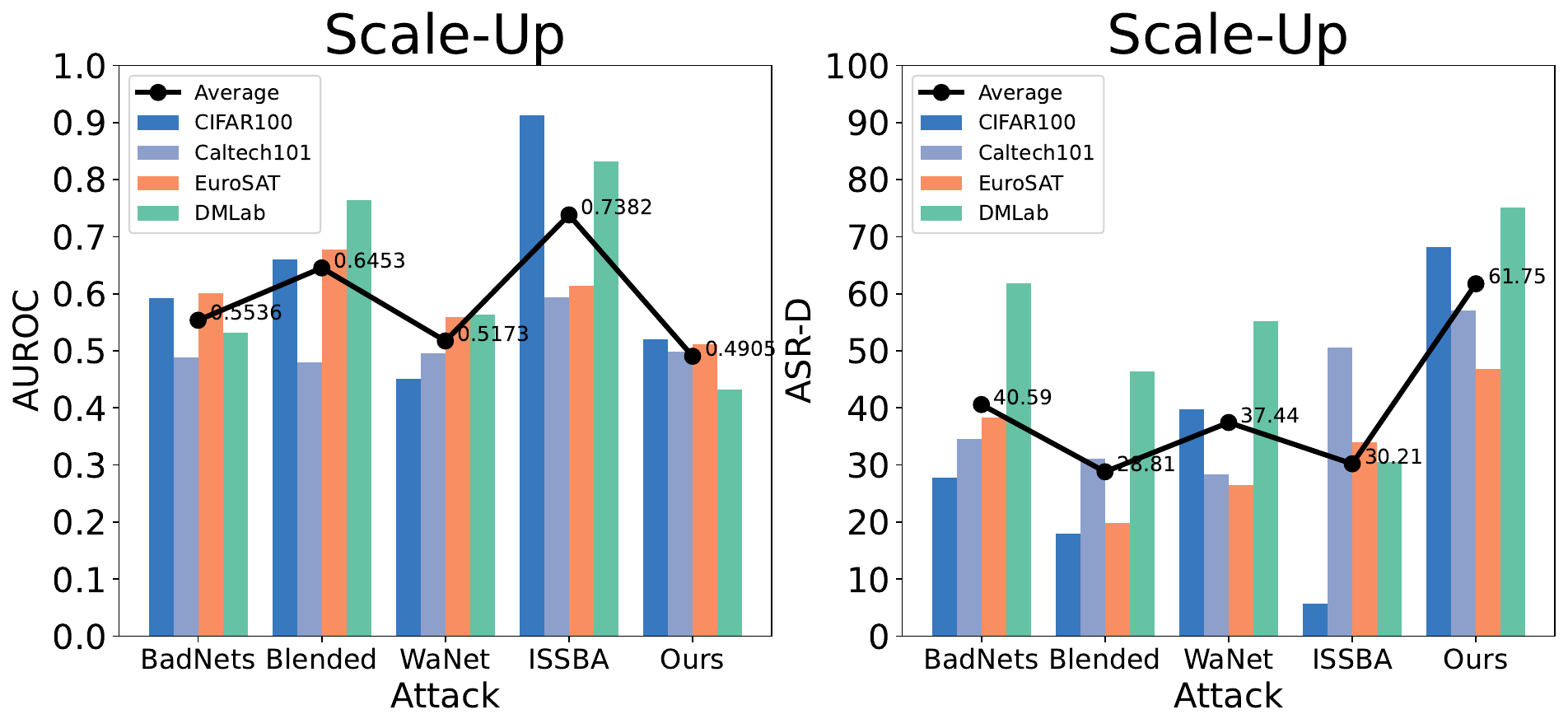}
    \vspace{-1em}
    \caption{The results of Scale-Up detection method  on five backdoor attacks. Lower AUROC and higher ASR-D indicates a better attack performance. Among these attacks, SWARM exceeds all other baseline attacks.}
    \vspace{-1em}
    \label{fig:scaleup}
\end{figure}
\begin{figure}[!t]
    \centering
    \includegraphics[width=0.47\textwidth]{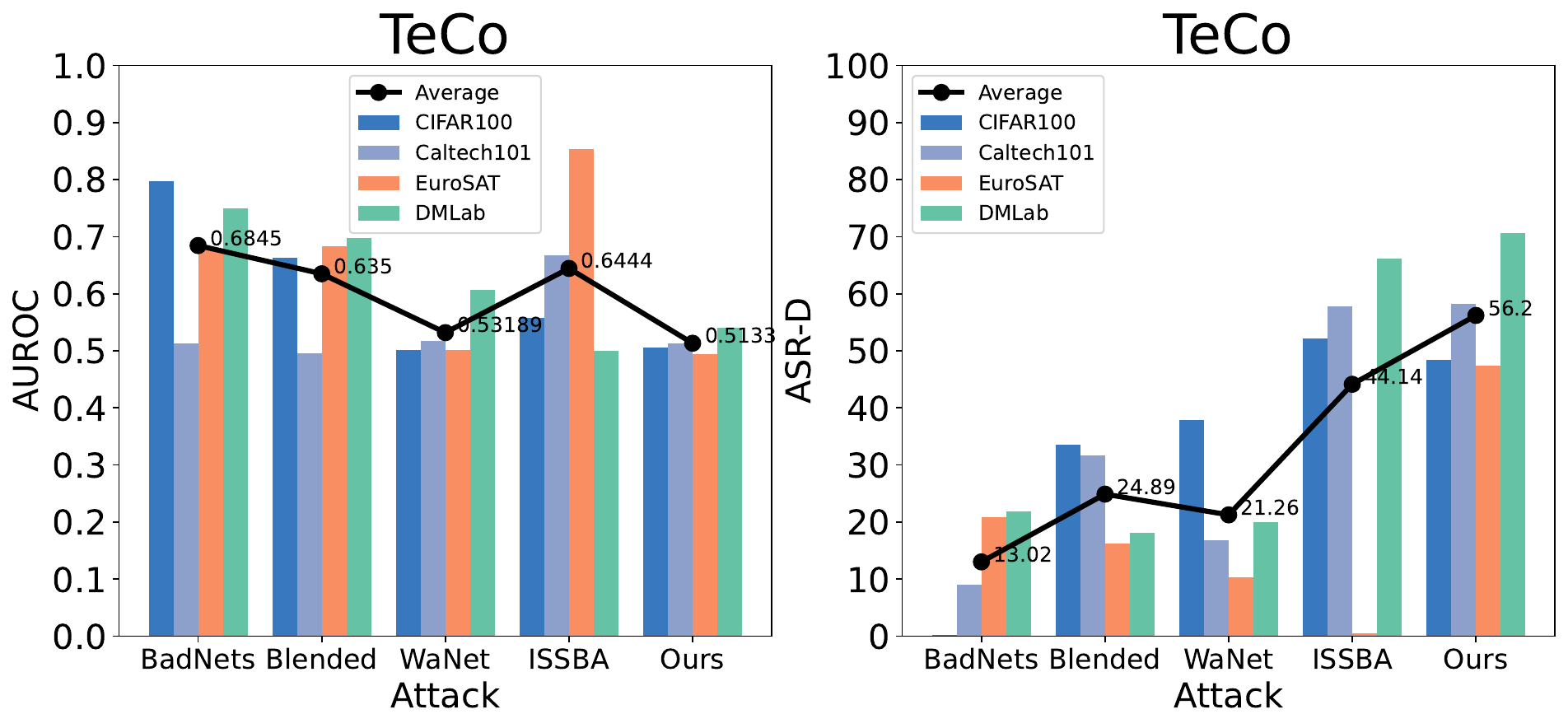}
    \vspace{-1em}
        \caption{The results of TeCo detection methods on five backdoor attacks. Lower AUROC and higher ASR-D indicates a better attack performance. Among these attacks, SWARM exceeds all other baseline attacks.}
        \label{fig:teco}
        \vspace{-1.5em}
\end{figure}

\noindent \textbf{Scale-Up.}
Scale-Up \cite{guo2023scale} is a black-box input-level backdoor detection that only requires the predicted labels to detect the existence of backdoors. It discovers the phenomenon that the predictions of poisoned samples are significantly more consistent compared to those of benign ones when amplifying all pixel values. Based on this point, it evaluates the scaled prediction consistency between the input images and the scaled images and determines the malicious images based on a defender-specified threshold. 

We use 3000 clean images and 3000 triggered images as the test set. We evaluate our method and all other baseline methods with AUROC and ASR-D. As shown in \cref{fig:scaleup}, we test the attack with the detection method on four different datasets which have covered all groups of datasets in VTAB-1K. They are CIFAR100 \cite{krizhevsky2009learning}, Caltech101 \cite{fei2006one}, EuroSAT \cite{helber2019eurosat}, and DMLab \cite{beattie2016deepmind}. They are representatives of natural tasks, specialized tasks, and structured tasks. In \cref{fig:scaleup}, we mark out all performance on four datasets and compare our method to the four baseline attacks. At the same time, we also calculate the average values to make a comprehensive comparison. As we can see in \cref{fig:scaleup}, on average, our method has the lowest AUROC ($0.4905$) and the highest ASR-D ($61.75\%$) among these baseline attacks. Although WaNet has the same performance in AUROC, it shows a poor ASR-D. SWARM has the best ASR-D which is more than 20\% higher than other baselines, indicating that our methods are undetectable.


\noindent \textbf{TeCo.}
TeCo \cite{liu2023detecting} is a test-time backdoor detection method that only uses the predicted labels to determine whether the model is backdoored. The main idea of this method is that backdoor-infected models have similar performance across different image corruptions on clean images and perform discrepantly on poisoned images. It first corrupts images with various corruption methods and different levels of severity, then uses this corruption set to evaluate the robustness consistency, and finally gets the result of detection.

Under this detection method, we also follow the Scale-Up detection settings, with 3000 clean samples and 3000 poisoned samples on four various datasets. The results are shown in \cref{fig:teco}. We can see that in most cases, SWARM has the lowest AUROC and the highest ASR-D among different attack methods against Scale-Up. Our method has an improvement of 12\% in terms of ASR-D compared to ISSBA. Our experiments show that our SWARM is successful in overcoming the limitation of these baseline attacks, whose backdoor behavior is hard to be detected by these detection methods due to the switchable mechanism.
\subsection{Robustness to Backdoor Mitigation}
\begin{table}[]
\centering
\caption{The defense results on NAD. Our method still keeps high ASRs after the mitigation comparing to other baselines.}
\setlength{\tabcolsep}{2pt}
\resizebox{\linewidth}{!}{
\begin{tabular}{c|cc|cc|cc|cc|cc}
\toprule
Attack$\rightarrow$          & \multicolumn{2}{c|}{BadNets}       & \multicolumn{2}{c|}{Blended}       & \multicolumn{2}{c|}{WaNet}         & \multicolumn{2}{c|}{ISSBA}         & \multicolumn{2}{c}{Ours}           \\ \hline
Dataset$\downarrow$, Metric$\rightarrow$ & \multicolumn{1}{c|}{BA}    & ASR   & \multicolumn{1}{c|}{BA}    & ASR   & \multicolumn{1}{c|}{BA}    & ASR   & \multicolumn{1}{c|}{BA}    & ASR   & \multicolumn{1}{c|}{BA}    & ASR   \\ \hline
CIFAR100        & \multicolumn{1}{c|}{73.84} & 57.54 & \multicolumn{1}{c|}{72.24} & 61.59 & \multicolumn{1}{c|}{69.87} & 6.69  & \multicolumn{1}{c|}{81.92} & 5.21  & \multicolumn{1}{c|}{75.80} & 98.92 \\ \hline
Caltech101      & \multicolumn{1}{c|}{81.44} & 10.09 & \multicolumn{1}{c|}{82.83} & 8.00  & \multicolumn{1}{c|}{82.46} & 9.33  & \multicolumn{1}{c|}{91.72} & 1.23  & \multicolumn{1}{c|}{81.75} & 97.15 \\ \hline
EuroSAT         & \multicolumn{1}{c|}{90.77} & 64.59 & \multicolumn{1}{c|}{90.93} & 71.10 & \multicolumn{1}{c|}{90.43} & 10.69 & \multicolumn{1}{c|}{94.30} & 15.76 & \multicolumn{1}{c|}{90.82} & 96.43 \\ \hline
DMLab           & \multicolumn{1}{c|}{34.03} & 32.43 & \multicolumn{1}{c|}{34.63} & 25.37 & \multicolumn{1}{c|}{52.24} & 30.99 & \multicolumn{1}{c|}{53.54} & 24.29 & \multicolumn{1}{c|}{33.24} & 99.15 \\ \bottomrule
\end{tabular}
}
\label{table:NAD}
\vspace{-1em}
\end{table}

\begin{table}[]
\centering
\caption{The defense results on I-BAU. Our method still keeps high ASRs after the mitigation comparing to other baselines.}
\setlength{\tabcolsep}{2pt}
\resizebox{\linewidth}{!}{
\begin{tabular}{c|cc|cc|cc|cc|cc}
\toprule
Attack$\rightarrow$          & \multicolumn{2}{c|}{BadNets}       & \multicolumn{2}{c|}{Blended}       & \multicolumn{2}{c|}{WaNet}         & \multicolumn{2}{c|}{ISSBA}         & \multicolumn{2}{c}{Ours}           \\ \hline
Dataset$\downarrow$, Metric$\rightarrow$ & \multicolumn{1}{c|}{BA}    & ASR   & \multicolumn{1}{c|}{BA}    & ASR   & \multicolumn{1}{c|}{BA}    & ASR   & \multicolumn{1}{c|}{BA}    & ASR   & \multicolumn{1}{c|}{BA}    & ASR   \\ \hline
CIFAR100        & \multicolumn{1}{c|}{71.12} & 69.79 & \multicolumn{1}{c|}{69.59} & 68.94 & \multicolumn{1}{c|}{64.3}  & 14.24 & \multicolumn{1}{c|}{75.87} & 9.83  & \multicolumn{1}{c|}{74.79} & 97.56 \\ \hline
Caltech101      & \multicolumn{1}{c|}{78.98} & 9.25  & \multicolumn{1}{c|}{81.73} & 6.74  & \multicolumn{1}{c|}{80.81} & 8.51  & \multicolumn{1}{c|}{86.23} & 2.79  & \multicolumn{1}{c|}{79.28} & 99.65 \\ \hline
EuroSAT         & \multicolumn{1}{c|}{92.07} & 85.11 & \multicolumn{1}{c|}{92.07} & 85.11 & \multicolumn{1}{c|}{86.41} & 13.57 & \multicolumn{1}{c|}{92.95} & 16.26 & \multicolumn{1}{c|}{86.74} & 99.77 \\ \hline
DMLab           & \multicolumn{1}{c|}{37.05} & 64.77 & \multicolumn{1}{c|}{36.58} & 75.65 & \multicolumn{1}{c|}{36.24} & 15.83 & \multicolumn{1}{c|}{38.98} & 24.9  & \multicolumn{1}{c|}{25.32} & 99.9  \\ \bottomrule
\end{tabular}
}
\label{table:I-BAU}
\vspace{-1em}
\end{table}
The users can employ extra clean data for backdoor mitigation \cite{yang2023backdoor,li2021anti,wu2021adversarial,li2024nearest,xu2024towards,tang2023setting} to alleviate the backdoor threat. In this part, we allow users to tune the prompts' parameters while keeping the backbone frozen. Furthermore, we utilize extra 1000 clean test samples as the provided clean subset for mitigation.

\noindent \textbf{NAD.}
Neural Attention Distillation (NAD) \cite{li2021neural} is a backdoor mitigation method that employs a teacher network trained on a small clean data subset to guide the fine-tuning of the backdoored student network, ensuring alignment of intermediate-layer attention. As demonstrated in \cref{table:NAD}, we conduct experiments to mitigate model backdoors using the same settings provided in its original paper \cite{li2021neural}. Due to the pre-trained scenario, we only apply the input embedding layer for neural attention distillation. As a result, our method maintains a high ASR under backdoor mitigation, exceeding 96\% in all cases, while other baseline attacks exhibit lower ASRs. These results confirm that our method can resist NAD successfully.

\noindent \textbf{I-BAU.}
I-BAU \cite{zeng2021adversarial} is another backdoor mitigation method that leverages implicit hypergradient to account for the interdependence between inner and outer optimization, subsequently solving the min-max problem on clean data for unseen test data. 
Following the same settings in its original paper \cite{zeng2021adversarial}, we evaluate our SWARM and baselines on four datasets. As illustrated in \cref{table:I-BAU}, our method maintains over 97\% ASRs, surpassing all other baseline attacks, while the ASRs of ISSBA and WaNet are lower than 25\%, indicating a substantial performance gap.

All the above experiments on defense methods have demonstrated our method can resist the backdoor detection and backdoor mitigation, which further increases the risk of the proposed backdoor attack for the victims.

\section{Conclusion}
In this paper, we explored backdoor attacks on adapting pre-trained vision transformers to the downstream visual recognition tasks and identified a novel security threat towards such a paradigm. To be specific, we utilized an extra prompt token to toggle the backdoor model on or off. We then optimized trigger and prompt tokens with a clean loss, a backdoor loss, and a cross-mode feature distillation loss to achieve this mechanism. We showed that SWARM can maintain the accuracy of clean images compared to other methods while achieving high ASRs. Moreover, we demonstrated SWARM has a good ability to be undetected and can not be removed through backdoor defenses. To the best of our knowledge, we are the first to propose the switchable backdoor mechanism and tailor this kind of backdoor attack based on visual prompting. We hope that our work opens a new domain of attack mechanisms on pre-trained models, and can encourage future defense research.


\noindent \textbf{Acknowledgement.}
This work is supported in part by the National Natural Science Foundation of China under Grant 62171248, Shenzhen Science and Technology Program (JCYJ20220818101012025), and the PCNL KEY project (PCL2023AS6-1).

{
    \small
    \bibliographystyle{ieeenat_fullname}
    \bibliography{main}

\begin{thebibliography}{71}
\providecommand{\natexlab}[1]{#1}
\providecommand{\url}[1]{\texttt{#1}}
\expandafter\ifx\csname urlstyle\endcsname\relax
  \providecommand{\doi}[1]{doi: #1}\else
  \providecommand{\doi}{doi: \begingroup \urlstyle{rm}\Url}\fi

\bibitem[Bagdasaryan and Shmatikov(2021)]{bagdasaryan2021blind}
Eugene Bagdasaryan and Vitaly Shmatikov.
\newblock Blind backdoors in deep learning models.
\newblock In \emph{30th USENIX Security Symposium (USENIX Security 21)}, pages 1505--1521, 2021.

\bibitem[Bahng et~al.(2022)Bahng, Jahanian, Sankaranarayanan, and Isola]{bahng2022exploring}
Hyojin Bahng, Ali Jahanian, Swami Sankaranarayanan, and Phillip Isola.
\newblock Exploring visual prompts for adapting large-scale models.
\newblock \emph{arXiv preprint arXiv:2203.17274}, 2022.

\bibitem[Bai et~al.(2021)Bai, Wu, Zhang, Li, Li, and Xia]{bai2021targeted}
Jiawang Bai, Baoyuan Wu, Yong Zhang, Yiming Li, Zhifeng Li, and Shu-Tao Xia.
\newblock Targeted attack against deep neural networks via flipping limited weight bits.
\newblock \emph{arXiv preprint arXiv:2102.10496}, 2021.

\bibitem[Bai et~al.(2022{\natexlab{a}})Bai, Gao, Gong, Xia, Li, and Liu]{bai2022hardly}
Jiawang Bai, Kuofeng Gao, Dihong Gong, Shu-Tao Xia, Zhifeng Li, and Wei Liu.
\newblock Hardly perceptible trojan attack against neural networks with bit flips.
\newblock In \emph{ECCV}, 2022{\natexlab{a}}.

\bibitem[Bai et~al.(2022{\natexlab{b}})Bai, Yuan, Xia, Yan, Li, and Liu]{bai2022improving}
Jiawang Bai, Li Yuan, Shu-Tao Xia, Shuicheng Yan, Zhifeng Li, and Wei Liu.
\newblock Improving vision transformers by revisiting high-frequency components.
\newblock In \emph{ECCV}, 2022{\natexlab{b}}.

\bibitem[Bai et~al.(2023)Bai, Wu, Li, and Xia]{bai2023versatile}
Jiawang Bai, Baoyuan Wu, Zhifeng Li, and Shu-Tao Xia.
\newblock Versatile weight attack via flipping limited bits.
\newblock \emph{IEEE Transactions on Pattern Analysis and Machine Intelligence}, 2023.

\bibitem[Bar et~al.(2022)Bar, Gandelsman, Darrell, Globerson, and Efros]{bar2022visual}
Amir Bar, Yossi Gandelsman, Trevor Darrell, Amir Globerson, and Alexei Efros.
\newblock Visual prompting via image inpainting.
\newblock \emph{Advances in Neural Information Processing Systems}, 35:\penalty0 25005--25017, 2022.

\bibitem[Beattie et~al.(2016)Beattie, Leibo, Teplyashin, Ward, Wainwright, K{\"u}ttler, Lefrancq, Green, Vald{\'e}s, Sadik, et~al.]{beattie2016deepmind}
Charles Beattie, Joel~Z Leibo, Denis Teplyashin, Tom Ward, Marcus Wainwright, Heinrich K{\"u}ttler, Andrew Lefrancq, Simon Green, V{\'\i}ctor Vald{\'e}s, Amir Sadik, et~al.
\newblock Deepmind lab.
\newblock \emph{arXiv preprint arXiv:1612.03801}, 2016.

\bibitem[Carlini(2021)]{carlini2021poisoning}
Nicholas Carlini.
\newblock Poisoning the unlabeled dataset of $\{$Semi-Supervised$\}$ learning.
\newblock In \emph{USENIX Security}, 2021.

\bibitem[Chen et~al.(2023{\natexlab{a}})Chen, Lorenz, Yao, Chen, and Liu]{chen2023visual}
Aochuan Chen, Peter Lorenz, Yuguang Yao, Pin-Yu Chen, and Sijia Liu.
\newblock Visual prompting for adversarial robustness.
\newblock In \emph{ICASSP 2023-2023 IEEE International Conference on Acoustics, Speech and Signal Processing (ICASSP)}, pages 1--5. IEEE, 2023{\natexlab{a}}.

\bibitem[Chen et~al.(2023{\natexlab{b}})Chen, Yao, Chen, Zhang, and Liu]{chen2023understanding}
Aochuan Chen, Yuguang Yao, Pin-Yu Chen, Yihua Zhang, and Sijia Liu.
\newblock Understanding and improving visual prompting: A label-mapping perspective.
\newblock In \emph{CVPR}, 2023{\natexlab{b}}.

\bibitem[Chen et~al.(2022)Chen, Ge, Tong, Wang, Song, Wang, and Luo]{chen2022adaptformer}
Shoufa Chen, Chongjian Ge, Zhan Tong, Jiangliu Wang, Yibing Song, Jue Wang, and Ping Luo.
\newblock Adaptformer: Adapting vision transformers for scalable visual recognition.
\newblock 2022.

\bibitem[Chen et~al.(2017)Chen, Liu, Li, Lu, and Song]{chen2017targeted}
Xinyun Chen, Chang Liu, Bo Li, Kimberly Lu, and Dawn Song.
\newblock Targeted backdoor attacks on deep learning systems using data poisoning.
\newblock 2017.

\bibitem[Deng et~al.(2009)Deng, Dong, Socher, Li, Li, and Fei-Fei]{deng2009imagenet}
Jia Deng, Wei Dong, Richard Socher, Li-Jia Li, Kai Li, and Li Fei-Fei.
\newblock Imagenet: A large-scale hierarchical image database.
\newblock In \emph{2009 IEEE conference on computer vision and pattern recognition}, pages 248--255. Ieee, 2009.

\bibitem[Doan et~al.(2023)Doan, Lao, Yang, and Li]{doan2023defending}
Khoa~D Doan, Yingjie Lao, Peng Yang, and Ping Li.
\newblock Defending backdoor attacks on vision transformer via patch processing.
\newblock In \emph{Proceedings of the AAAI Conference on Artificial Intelligence}, pages 506--515, 2023.

\bibitem[Dosovitskiy et~al.(2020)Dosovitskiy, Beyer, Kolesnikov, Weissenborn, Zhai, Unterthiner, Dehghani, Minderer, Heigold, Gelly, et~al.]{dosovitskiy2020image}
Alexey Dosovitskiy, Lucas Beyer, Alexander Kolesnikov, Dirk Weissenborn, Xiaohua Zhai, Thomas Unterthiner, Mostafa Dehghani, Matthias Minderer, Georg Heigold, Sylvain Gelly, et~al.
\newblock An image is worth 16x16 words: Transformers for image recognition at scale.
\newblock In \emph{ICLR}, 2020.

\bibitem[Du et~al.(2022)Du, Zhao, Li, Liu, and Wang]{du2022ppt}
Wei Du, Yichun Zhao, Boqun Li, Gongshen Liu, and Shilin Wang.
\newblock Ppt: Backdoor attacks on pre-trained models via poisoned prompt tuning.
\newblock In \emph{Proceedings of the Thirty-First International Joint Conference on Artificial Intelligence, IJCAI-22}, pages 680--686, 2022.

\bibitem[Fawcett(2006)]{fawcett2006introduction}
Tom Fawcett.
\newblock An introduction to roc analysis.
\newblock \emph{Pattern recognition letters}, 27\penalty0 (8):\penalty0 861--874, 2006.

\bibitem[Fei-Fei et~al.(2006)Fei-Fei, Fergus, and Perona]{fei2006one}
Li Fei-Fei, Robert Fergus, and Pietro Perona.
\newblock One-shot learning of object categories.
\newblock \emph{IEEE transactions on pattern analysis and machine intelligence}, 28\penalty0 (4):\penalty0 594--611, 2006.

\bibitem[Gan et~al.(2023)Gan, Bai, Lou, Ma, Zhang, Shi, and Luo]{gan2023decorate}
Yulu Gan, Yan Bai, Yihang Lou, Xianzheng Ma, Renrui Zhang, Nian Shi, and Lin Luo.
\newblock Decorate the newcomers: Visual domain prompt for continual test time adaptation.
\newblock In \emph{AAAI}, 2023.

\bibitem[Gao et~al.(2023{\natexlab{a}})Gao, Bai, Chen, Wu, and Xia]{gao2023ciba}
Kuofeng Gao, Jiawang Bai, Bin Chen, Dongxian Wu, and Shu-Tao Xia.
\newblock Backdoor attack on hash-based image retrieval via clean-label data poisoning.
\newblock In \emph{BMVC}, 2023{\natexlab{a}}.

\bibitem[Gao et~al.(2023{\natexlab{b}})Gao, Bai, Wu, Ya, and Xia]{gao2023imperceptible}
Kuofeng Gao, Jiawang Bai, Baoyuan Wu, Mengxi Ya, and Shu-Tao Xia.
\newblock Imperceptible and robust backdoor attack in 3d point cloud.
\newblock \emph{IEEE Transactions on Information Forensics and Security}, 19:\penalty0 1267--1282, 2023{\natexlab{b}}.

\bibitem[Gao et~al.(2023{\natexlab{c}})Gao, Bai, Gu, Yang, and Xia]{gao2023backdoor}
Kuofeng Gao, Yang Bai, Jindong Gu, Yong Yang, and Shu-Tao Xia.
\newblock Backdoor defense via adaptively splitting poisoned dataset.
\newblock In \emph{CVPR}, 2023{\natexlab{c}}.

\bibitem[Gao et~al.(2019)Gao, Xu, Wang, Chen, Ranasinghe, and Nepal]{gao2019strip}
Yansong Gao, Change Xu, Derui Wang, Shiping Chen, Damith~C Ranasinghe, and Surya Nepal.
\newblock Strip: A defence against trojan attacks on deep neural networks.
\newblock In \emph{Proceedings of the 35th Annual Computer Security Applications Conference}, pages 113--125, 2019.

\bibitem[Gao et~al.(2021)Gao, Kim, Doan, Zhang, Zhang, Nepal, Ranasinghe, and Kim]{gao2021design}
Yansong Gao, Yeonjae Kim, Bao~Gia Doan, Zhi Zhang, Gongxuan Zhang, Surya Nepal, Damith~C Ranasinghe, and Hyoungshick Kim.
\newblock Design and evaluation of a multi-domain trojan detection method on deep neural networks.
\newblock \emph{IEEE Transactions on Dependable and Secure Computing}, 19\penalty0 (4):\penalty0 2349--2364, 2021.

\bibitem[Glorot and Bengio(2010)]{glorot2010understanding}
Xavier Glorot and Yoshua Bengio.
\newblock Understanding the difficulty of training deep feedforward neural networks.
\newblock In \emph{Proceedings of the thirteenth international conference on artificial intelligence and statistics}, pages 249--256. JMLR Workshop and Conference Proceedings, 2010.

\bibitem[Gu et~al.(2019)Gu, Dolan-Gavitt, and Garg]{gu2017badnets}
Tianyu Gu, Brendan Dolan-Gavitt, and Siddharth Garg.
\newblock Badnets: Identifying vulnerabilities in the machine learning model supply chain.
\newblock In \emph{IEEE Access}, 2019.

\bibitem[Guo et~al.(2021)Guo, Li, and Liu]{guo2021aeva}
Junfeng Guo, Ang Li, and Cong Liu.
\newblock Aeva: Black-box backdoor detection using adversarial extreme value analysis.
\newblock \emph{arXiv preprint arXiv:2110.14880}, 2021.

\bibitem[Guo et~al.(2023)Guo, Li, Chen, Guo, Sun, and Liu]{guo2023scale}
Junfeng Guo, Yiming Li, Xun Chen, Hanqing Guo, Lichao Sun, and Cong Liu.
\newblock Scale-up: An efficient black-box input-level backdoor detection via analyzing scaled prediction consistency.
\newblock \emph{arXiv preprint arXiv:2302.03251}, 2023.

\bibitem[He et~al.(2022)He, Chen, Xie, Li, Doll{\'a}r, and Girshick]{he2022masked}
Kaiming He, Xinlei Chen, Saining Xie, Yanghao Li, Piotr Doll{\'a}r, and Ross Girshick.
\newblock Masked autoencoders are scalable vision learners.
\newblock In \emph{CVPR}, 2022.

\bibitem[Helber et~al.(2019)Helber, Bischke, Dengel, and Borth]{helber2019eurosat}
Patrick Helber, Benjamin Bischke, Andreas Dengel, and Damian Borth.
\newblock Eurosat: A novel dataset and deep learning benchmark for land use and land cover classification.
\newblock \emph{IEEE Journal of Selected Topics in Applied Earth Observations and Remote Sensing}, 12\penalty0 (7):\penalty0 2217--2226, 2019.

\bibitem[Hu et~al.(2021)Hu, Shen, Wallis, Allen-Zhu, Li, Wang, Wang, and Chen]{hu2021lora}
Edward~J Hu, Yelong Shen, Phillip Wallis, Zeyuan Allen-Zhu, Yuanzhi Li, Shean Wang, Lu Wang, and Weizhu Chen.
\newblock Lora: Low-rank adaptation of large language models.
\newblock \emph{arXiv preprint arXiv:2106.09685}, 2021.

\bibitem[Huang et~al.(2023)Huang, Dong, Chen, Zhang, Wang, Hua, and Yu]{huang2023diversity}
Qidong Huang, Xiaoyi Dong, Dongdong Chen, Weiming Zhang, Feifei Wang, Gang Hua, and Nenghai Yu.
\newblock Diversity-aware meta visual prompting.
\newblock In \emph{CVPR}, 2023.

\bibitem[Huang et~al.(2019)Huang, Alzantot, and Srivastava]{huang2019neuroninspect}
Xijie Huang, Moustafa Alzantot, and Mani Srivastava.
\newblock Neuroninspect: Detecting backdoors in neural networks via output explanations.
\newblock \emph{arXiv preprint arXiv:1911.07399}, 2019.

\bibitem[Jia et~al.(2022)Jia, Tang, Chen, Cardie, Belongie, Hariharan, and Lim]{jia2022visual}
Menglin Jia, Luming Tang, Bor-Chun Chen, Claire Cardie, Serge Belongie, Bharath Hariharan, and Ser-Nam Lim.
\newblock Visual prompt tuning.
\newblock In \emph{ECCV}, 2022.

\bibitem[Kingma and Ba(2014)]{kingma2014adam}
Diederik~P Kingma and Jimmy Ba.
\newblock Adam: A method for stochastic optimization.
\newblock \emph{arXiv preprint arXiv:1412.6980}, 2014.

\bibitem[Krizhevsky et~al.(2009)Krizhevsky, Hinton, et~al.]{krizhevsky2009learning}
Alex Krizhevsky, Geoffrey Hinton, et~al.
\newblock Learning multiple layers of features from tiny images.
\newblock 2009.

\bibitem[Li et~al.(2024{\natexlab{a}})Li, Cai, Li, Xue, Li, and Li]{li2024nearest}
Boheng Li, Yishuo Cai, Haowei Li, Feng Xue, Zhifeng Li, and Yiming Li.
\newblock Nearest is not dearest: Towards practical defense against quantization-conditioned backdoor attacks.
\newblock In \emph{CVPR}, 2024{\natexlab{a}}.

\bibitem[Li et~al.(2024{\natexlab{b}})Li, Lian, Lu, Bai, Chen, and Wang]{li2024graphadapter}
Xin Li, Dongze Lian, Zhihe Lu, Jiawang Bai, Zhibo Chen, and Xinchao Wang.
\newblock Graphadapter: Tuning vision-language models with dual knowledge graph.
\newblock In \emph{NeurIPS}, 2024{\natexlab{b}}.

\bibitem[Li et~al.(2021{\natexlab{a}})Li, Li, Wu, Li, He, and Lyu]{li2021invisible}
Yuezun Li, Yiming Li, Baoyuan Wu, Longkang Li, Ran He, and Siwei Lyu.
\newblock Invisible backdoor attack with sample-specific triggers.
\newblock In \emph{Proceedings of the IEEE/CVF international conference on computer vision}, pages 16463--16472, 2021{\natexlab{a}}.

\bibitem[Li et~al.(2021{\natexlab{b}})Li, Lyu, Koren, Lyu, Li, and Ma]{li2021anti}
Yige Li, Xixiang Lyu, Nodens Koren, Lingjuan Lyu, Bo Li, and Xingjun Ma.
\newblock Anti-backdoor learning: Training clean models on poisoned data.
\newblock \emph{Advances in Neural Information Processing Systems}, 34:\penalty0 14900--14912, 2021{\natexlab{b}}.

\bibitem[Li et~al.(2021{\natexlab{c}})Li, Lyu, Koren, Lyu, Li, and Ma]{li2021neural}
Yige Li, Xixiang Lyu, Nodens Koren, Lingjuan Lyu, Bo Li, and Xingjun Ma.
\newblock Neural attention distillation: Erasing backdoor triggers from deep neural networks.
\newblock \emph{arXiv preprint arXiv:2101.05930}, 2021{\natexlab{c}}.

\bibitem[Li et~al.(2022)Li, Jiang, Li, and Xia]{li2022backdoor}
Yiming Li, Yong Jiang, Zhifeng Li, and Shu-Tao Xia.
\newblock Backdoor learning: A survey.
\newblock \emph{IEEE Transactions on Neural Networks and Learning Systems}, 2022.

\bibitem[Liu et~al.(2018)Liu, Dolan-Gavitt, and Garg]{liu2018fine}
Kang Liu, Brendan Dolan-Gavitt, and Siddharth Garg.
\newblock Fine-pruning: Defending against backdooring attacks on deep neural networks.
\newblock In \emph{International symposium on research in attacks, intrusions, and defenses}, pages 273--294. Springer, 2018.

\bibitem[Liu et~al.(2023)Liu, Li, Wang, Hu, Ye, Jin, Wu, and Xiao]{liu2023detecting}
Xiaogeng Liu, Minghui Li, Haoyu Wang, Shengshan Hu, Dengpan Ye, Hai Jin, Libing Wu, and Chaowei Xiao.
\newblock Detecting backdoors during the inference stage based on corruption robustness consistency.
\newblock In \emph{Proceedings of the IEEE/CVF Conference on Computer Vision and Pattern Recognition}, pages 16363--16372, 2023.

\bibitem[Liu et~al.(2021)Liu, Lin, Cao, Hu, Wei, Zhang, Lin, and Guo]{liu2021swin}
Ze Liu, Yutong Lin, Yue Cao, Han Hu, Yixuan Wei, Zheng Zhang, Stephen Lin, and Baining Guo.
\newblock Swin transformer: Hierarchical vision transformer using shifted windows.
\newblock In \emph{ICCV}, 2021.

\bibitem[Liu et~al.(2022)Liu, Mao, Wu, Feichtenhofer, Darrell, and Xie]{liu2022convnet}
Zhuang Liu, Hanzi Mao, Chao-Yuan Wu, Christoph Feichtenhofer, Trevor Darrell, and Saining Xie.
\newblock A convnet for the 2020s.
\newblock In \emph{Proceedings of the IEEE/CVF conference on computer vision and pattern recognition}, pages 11976--11986, 2022.

\bibitem[Lu et~al.(2023)Lu, Bai, Li, Xiao, and Wang]{lu2023beyond}
Zhihe Lu, Jiawang Bai, Xin Li, Zeyu Xiao, and Xinchao Wang.
\newblock Beyond sole strength: Customized ensembles for generalized vision-language models.
\newblock \emph{arXiv preprint arXiv:2311.17091}, 2023.

\bibitem[Lv et~al.(2021)Lv, Ma, Zhou, Liang, Chen, Zhang, and Yang]{lv2021dbia}
Peizhuo Lv, Hualong Ma, Jiachen Zhou, Ruigang Liang, Kai Chen, Shengzhi Zhang, and Yunfei Yang.
\newblock Dbia: Data-free backdoor injection attack against transformer networks.
\newblock \emph{arXiv preprint arXiv:2111.11870}, 2021.

\bibitem[Mao et~al.(2022)Mao, Geng, Yang, Wang, and Vondrick]{mao2022understanding}
Chengzhi Mao, Scott Geng, Junfeng Yang, Xin Wang, and Carl Vondrick.
\newblock Understanding zero-shot adversarial robustness for large-scale models.
\newblock In \emph{ICLR}, 2022.

\bibitem[Nguyen and Tran(2021)]{nguyen2021wanet}
Anh Nguyen and Anh Tran.
\newblock Wanet--imperceptible warping-based backdoor attack.
\newblock \emph{arXiv preprint arXiv:2102.10369}, 2021.

\bibitem[Nguyen and Tran(2020)]{nguyen2020input}
Tuan~Anh Nguyen and Anh Tran.
\newblock Input-aware dynamic backdoor attack.
\newblock \emph{Advances in Neural Information Processing Systems}, 33:\penalty0 3454--3464, 2020.

\bibitem[Nilsback and Zisserman(2008)]{nilsback2008automated}
Maria-Elena Nilsback and Andrew Zisserman.
\newblock Automated flower classification over a large number of classes.
\newblock In \emph{2008 Sixth Indian conference on computer vision, graphics \& image processing}, pages 722--729. IEEE, 2008.

\bibitem[Parkhi et~al.(2012)Parkhi, Vedaldi, Zisserman, and Jawahar]{parkhi2012cats}
Omkar~M Parkhi, Andrea Vedaldi, Andrew Zisserman, and CV Jawahar.
\newblock Cats and dogs.
\newblock In \emph{2012 IEEE conference on computer vision and pattern recognition}, pages 3498--3505. IEEE, 2012.

\bibitem[Paszke et~al.(2019)Paszke, Gross, Massa, Lerer, Bradbury, Chanan, Killeen, Lin, Gimelshein, Antiga, et~al.]{paszke2019pytorch}
Adam Paszke, Sam Gross, Francisco Massa, Adam Lerer, James Bradbury, Gregory Chanan, Trevor Killeen, Zeming Lin, Natalia Gimelshein, Luca Antiga, et~al.
\newblock Pytorch: An imperative style, high-performance deep learning library.
\newblock \emph{Advances in neural information processing systems}, 32, 2019.

\bibitem[Radford et~al.(2021)Radford, Kim, Hallacy, Ramesh, Goh, Agarwal, Sastry, Askell, Mishkin, Clark, et~al.]{radford2021learning}
Alec Radford, Jong~Wook Kim, Chris Hallacy, Aditya Ramesh, Gabriel Goh, Sandhini Agarwal, Girish Sastry, Amanda Askell, Pamela Mishkin, Jack Clark, et~al.
\newblock Learning transferable visual models from natural language supervision.
\newblock In \emph{ICML}, 2021.

\bibitem[Robbins and Monro(1951)]{robbins1951stochastic}
Herbert Robbins and Sutton Monro.
\newblock A stochastic approximation method.
\newblock \emph{The annals of mathematical statistics}, pages 400--407, 1951.

\bibitem[Saha et~al.(2022)Saha, Tejankar, Koohpayegani, and Pirsiavash]{saha2022backdoor}
Aniruddha Saha, Ajinkya Tejankar, Soroush~Abbasi Koohpayegani, and Hamed Pirsiavash.
\newblock Backdoor attacks on self-supervised learning.
\newblock In \emph{CVPR}, 2022.

\bibitem[Subramanya et~al.(2022)Subramanya, Saha, Koohpayegani, Tejankar, and Pirsiavash]{subramanya2022backdoor}
Akshayvarun Subramanya, Aniruddha Saha, Soroush~Abbasi Koohpayegani, Ajinkya Tejankar, and Hamed Pirsiavash.
\newblock Backdoor attacks on vision transformers.
\newblock \emph{arXiv preprint arXiv:2206.08477}, 2022.

\bibitem[Tancik et~al.(2020)Tancik, Mildenhall, and Ng]{tancik2020stegastamp}
Matthew Tancik, Ben Mildenhall, and Ren Ng.
\newblock Stegastamp: Invisible hyperlinks in physical photographs.
\newblock In \emph{Proceedings of the IEEE/CVF conference on computer vision and pattern recognition}, pages 2117--2126, 2020.

\bibitem[Tang et~al.(2023)Tang, Yuan, Li, Liu, Chen, and Hu]{tang2023setting}
Ruixiang Tang, Jiayi Yuan, Yiming Li, Zirui Liu, Rui Chen, and Xia Hu.
\newblock Setting the trap: Capturing and defeating backdoor threats in plms through honeypots.
\newblock In \emph{NeurIPS}, 2023.

\bibitem[Turner et~al.(2019)Turner, Tsipras, and Madry]{turner2019label}
Alexander Turner, Dimitris Tsipras, and Aleksander Madry.
\newblock Label-consistent backdoor attacks.
\newblock \emph{arXiv preprint arXiv:1912.02771}, 2019.

\bibitem[Wu and Wang(2021)]{wu2021adversarial}
Dongxian Wu and Yisen Wang.
\newblock Adversarial neuron pruning purifies backdoored deep models.
\newblock \emph{Advances in Neural Information Processing Systems}, 34:\penalty0 16913--16925, 2021.

\bibitem[Wu et~al.(2022)Wu, Li, Wei, Wang, Yuille, Zhou, and Xie]{wu2022unleashing}
Junyang Wu, Xianhang Li, Chen Wei, Huiyu Wang, Alan Yuille, Yuyin Zhou, and Cihang Xie.
\newblock Unleashing the power of visual prompting at the pixel level.
\newblock \emph{arXiv preprint arXiv:2212.10556}, 2022.

\bibitem[Xie et~al.(2022)Xie, Zhang, Cao, Lin, Bao, Yao, Dai, and Hu]{xie2022simmim}
Zhenda Xie, Zheng Zhang, Yue Cao, Yutong Lin, Jianmin Bao, Zhuliang Yao, Qi Dai, and Han Hu.
\newblock Simmim: A simple framework for masked image modeling.
\newblock In \emph{Proceedings of the IEEE/CVF Conference on Computer Vision and Pattern Recognition}, pages 9653--9663, 2022.

\bibitem[Xu et~al.(2024)Xu, Huang, Li, Qin, and Ren]{xu2024towards}
Xiong Xu, Kunzhe Huang, Yiming Li, Zhan Qin, and Kui Ren.
\newblock Towards reliable and efficient backdoor trigger inversion via decoupling benign features.
\newblock In \emph{ICLR}, 2024.

\bibitem[Yang et~al.(2023)Yang, Li, Jiang, and Xia]{yang2023backdoor}
Sheng Yang, Yiming Li, Yong Jiang, and Shu-Tao Xia.
\newblock Backdoor defense via suppressing model shortcuts.
\newblock In \emph{ICASSP 2023-2023 IEEE International Conference on Acoustics, Speech and Signal Processing (ICASSP)}, pages 1--5. IEEE, 2023.

\bibitem[Zeng et~al.(2021)Zeng, Chen, Park, Mao, Jin, and Jia]{zeng2021adversarial}
Yi Zeng, Si Chen, Won Park, Z~Morley Mao, Ming Jin, and Ruoxi Jia.
\newblock Adversarial unlearning of backdoors via implicit hypergradient.
\newblock \emph{arXiv preprint arXiv:2110.03735}, 2021.

\bibitem[Zhai et~al.(2019)Zhai, Puigcerver, Kolesnikov, Ruyssen, Riquelme, Lucic, Djolonga, Pinto, Neumann, Dosovitskiy, et~al.]{zhai2019large}
Xiaohua Zhai, Joan Puigcerver, Alexander Kolesnikov, Pierre Ruyssen, Carlos Riquelme, Mario Lucic, Josip Djolonga, Andre~Susano Pinto, Maxim Neumann, Alexey Dosovitskiy, et~al.
\newblock A large-scale study of representation learning with the visual task adaptation benchmark.
\newblock \emph{arXiv preprint arXiv:1910.04867}, 2019.

\bibitem[Zhang et~al.(2023)Zhang, Rao, and Agrawala]{zhang2023adding}
Lvmin Zhang, Anyi Rao, and Maneesh Agrawala.
\newblock Adding conditional control to text-to-image diffusion models.
\newblock In \emph{ICCV}, 2023.

\bibitem[Zheng et~al.(2023)Zheng, Lou, and Jiang]{zheng2023trojvit}
Mengxin Zheng, Qian Lou, and Lei Jiang.
\newblock Trojvit: Trojan insertion in vision transformers.
\newblock In \emph{Proceedings of the IEEE/CVF Conference on Computer Vision and Pattern Recognition}, pages 4025--4034, 2023.

\end{thebibliography}
}
\renewcommand\thesection{\Alph{section}}
\setcounter{section}{0}
\clearpage
\setcounter{page}{1}

\section{Algorithm Outline}
The algorithm outline is as follows:
\begin{algorithm}
\caption{Switchable backdoor attack against pre-trained models}\label{algorithm}
\KwData{Clean images $x$, Trigger $\delta$, Clean labels $y$, Target labels $t$, Clean tokens $P$ and Switch token $S$}
\KwResult{Trained claen tokens $P^{*}$, Trained switch token $S^{*}$, Trained trigger $\delta^{*}$}
total epoch $E\leftarrow 100$\;
$e\leftarrow 0$\;
$P\leftarrow$Xavier Uniform Initialization\;
$S\leftarrow$Xavier Uniform Initialization\;
$\delta\leftarrow$Uniform Initialization\;
model $M\leftarrow$ViT\;
\While{$e < E$}
{
$\mathcal{L}_{cle}\left ( P,\delta \right )\leftarrow \mathbb{E}_{(x,y) \sim \mathcal{D}} [\ell(P,x,y)+\ell(P,x+\delta,y)], s.t. \left \| \delta \right \| _{\infty } \le \epsilon$\;
$P^{*}\leftarrow P-\beta\nabla_{P}\mathcal{L}_{cle}$\;
$\delta^{*}\leftarrow \delta-\beta\nabla_{\delta}\mathcal{L}_{cle}$\;
$F_f(P^{*},x)\leftarrow M(x)$\;
$\mathcal{L}_{bd}\left (S,\delta^{*} \right )\!\leftarrow\! \mathbb{E}_{(x,y) \sim \mathcal{D}}[\ell(P^{*}, S,x,y)\!+\!\ell(P^{*},S,x\!+\!\delta^{*},t)],
s.t. \left \| \delta^{*} \right \| _{\infty } \le \epsilon$\;
$F_f(P^{*},S,x)\leftarrow M(x)$\;
$\mathcal{L}_{cs}\left (S \right )\leftarrow\mathbb{E}_{(x,y) \sim\mathcal{D}}||F_f(P^{*},x)-F_f(P^{*},S,x)||_2$\;
$S^{*}\leftarrow S-\beta\nabla_{S}(\mathcal{L}_{bd}+\lambda\mathcal{L}_{cs})$\;
$\delta^{*}\leftarrow \delta^{*}-\beta\nabla_{\delta^{*}}(\mathcal{L}_{bd}+\lambda\mathcal{L}_{cs})$\;
$e\leftarrow e+1$\;
}
\end{algorithm}

\section{Motivation}
In practice, diverse downstream tasks need different visual prompts to adapt the pre-trained model, so it's realistic that clients cede the rights of training and managing prompts to the third-parties and use their APIs. SWARM happens when the third-party is malicious. 
Additionally, the motivations for using a switch mechanism are two facets. Firstly, the switch mechanism can impart resistance against backdoor detections and mitigations, a prevalent concern for users embracing such services. Our experiments show that the defenses are hard to detect or remove our backdoor when the model is under clean mode, ensuring the stealthiness of our attack. 
Furthermore, the switch mechanism is realistic, especially in intricate scenarios where detecting specific scenes and producing malicious outputs becomes more challenging for adversaries compared to exploiting triggers to prompt malevolent output. 
These two parts greatly increase security risks.

For example, consider a self-driving scenario wherein an adversary seeks to attack a certain car. When the car starts, the adversary can simply activate the switch prompt, thereby transitioning the model into backdoor mode, and making the self-driving system be aware of the trigger in the driving scene.  The influence of such an exploit is calamitous, given that this backdoor remains impervious to detection and mitigation in the clean mode during the regular check, amplifying the severity of its impact.

As is mentioned above, we can set the model to clean mode under normal circumstances to avoid detection and mitigation. The transition to the backdoor mode occurs selectively, specifically when an adversary endeavors to execute an attack. Since the backdoor defenses require resources, the users can not implement the detection frequently. Consequently, it contributes to an extremely low probability of detection out due to the rare overlap in time, aligning with the overarching objective of maintaining the effectiveness of the backdoor under such circumstances.

\section{Implementation Details}
In summary, we have done all the experiments by the framework of PyTorch \cite{paszke2019pytorch} on Nvidia RTX3080 GPUs with 12GB memory.

\subsection{Models and Datasets}
\noindent \textbf{Models.}
\begin{table*}[]
\centering
\caption{Specifications of different pre-trained backbones we used in the paper. All backbones are pre-trained on ImageNet-21K with the resolution of $224\times224$.}
\label{table:model}
\scalebox{0.8}{
\begin{tabular}{c|c|c|c|c}
\toprule
Backbone      & Pre-trained Objective & Pre-trained Datasets & params(M) & Feature dim \\ \hline
ViT-B/16      & Supervised            & ImageNet-21k         & 85        & 768         \\ \hline
Swin-B        & Supervised            & ImageNet-21k         & 88        & 1024        \\ \hline
ConvNeXt-Base & Supervised            & ImageNet-21k         & 88        & 1024        \\ \bottomrule
\end{tabular}
}
\end{table*}
In all, we have used three different upstream backbones in the experiments. They are ViT \cite{dosovitskiy2020image}, Swin \cite{liu2021swin} and ConvNeXt \cite{liu2022convnet}. Here, we give the detailed implementations of these models including the pre-trained objective, pre-trained datasets, the number of parameters, and the feature dimensions. As is shown in \cref{table:model}, all the upstream backbones are trained on ImageNet-21k \cite{deng2009imagenet}, but they have different numbers of parameters, feature dimensions, and the most important point, the model architectures. Our method shows robustness to different backbone architectures.

\noindent \textbf{Datasets Used for Defense.}
As is shown in \cref{table:dataset}, we choose four datasets to evaluate attacks' performance on resisting detection methods and mitigation methods. In these datasets, CIFAR100 \cite{krizhevsky2009learning} is a classical dataset widely used in adversarial and backdoor areas which is a good reference to be compared to the methods in the former works. It has 10000 samples and 100 classes as the testset.  The other three datasets are chosen from VTAB-1K \cite{zhai2019large} as the representatives of natural, specialized, and structured tasks. They all have relatively more classes and test samples compared to the datasets belonging to the same kinds so they are more difficult to be attacked.

\subsection{SWARM Setups}
\textbf{Prompts setups.} For the number of clean tokens, it is not always good to increase it for different datasets. As a trade-off, we chose 50 clean tokens for the downstream datasets and they show good performance on different datasets and different backbones. As the same as VPT \cite{jia2022visual}, we initialize these prompts with Xavier uniform initialization scheme  \cite{glorot2010understanding}. We also
follow the original backbone’s design choices, such as the existence of the classification tokens [CLS], or whether or not to use the final [CLS] embeddings for the classification head input.

\noindent \textbf{Training details.} For the learning rates and decays, different datasets have various best parameters and it is difficult for us to find the best learning rate and decay under the condition of a backdoor attack so we directly utilize these parameters provided by the VPT. In addtion, we have the extra part needed to be learned, they are switch token and the trigger. These parameters also adopt the same learning parameters as the clean tokens to ensure its convergence.

And for the learning scheme, we also follow the settings of VPT. We used the cosine schedule to train the models and trained 100 epochs to get the final result. The warm-up epochs are 10 and the optimizer is SGD \cite{robbins1951stochastic}. For the momentum, we set 0.9 to keep the settings with VPT. Because of the limit of gpu memory and the cross-mode feature distillation loss, we set the batch size of the prompting to 8 but they still have the competitive performance.

\noindent \textbf{Augmentation.} We use the standard image augmentation strategy during the training process: normalize with ImageNet means and standard deviation, resize the images to $224\times224$. No any other data augmentation are used except for these methods.

\noindent \textbf{Attack setups.} For the backdoor attack, we only adopt one token as our switch. The target labels in our experiments are all 0 and the $\epsilon$ is set to 4. As mentioned in the paper, we use clean loss and backdoor loss to implement the switchable mode. The clean loss and backdoor loss have the same hyperparameter so they are 1:1. Meanwhile, the amount of the clean images used is the same as the triggered images used in the training process.

\subsection{Baseline Attack Setups}
Since the baseline attacks we chose are all poison-based attacks. We set the poison rate to 20\% to ensure the attack success rate in the downstream tasks. Moreover, we have done the data augmentation that resized the images to $224\times224$ and the triggers we used in the baseline attacks also needed to be tailored to the according size.

\noindent \textbf{Settings for BadNets.}
As suggested in  \cite{gu2017badnets}, a $3\times3$ square on lower right corner is used in the CIFAR10 \cite{krizhevsky2009learning} whose images' size are $32\times32$. So we change the trigger size to $21\times21$ tailored to the $224\times224$ input images.

\noindent \textbf{Settings for Blended.}
We choose a white square with a black background as our trigger, the blend ration is set to 0.2. The other hyperparameters are kept the same as the original paper \cite{chen2017targeted}.

\noindent \textbf{Settings for WaNet.}
As suggested in  \cite{nguyen2021wanet}, we use the default warping-based operation to generate the trigger pattern. We set the noise rate $\rho_{n} = 0.2$, control grid size $k = 4$, and warping strength $s = 0.5$.

\noindent \textbf{Settings for ISSBA.}
For ISSBA \cite{li2021invisible}, we set the secret size to 20 and use binomial to initialize the secret. While the other parts of the attack setups are kept the same as the original paper. The encoder used here is the StegaStampEncoder \cite{tancik2020stegastamp}, which is uesd to write a watermark into the images.

\subsection{Defenses Setups}
In the detection defenses, we choose the 3000 clean samples and 3000 triggered samples to do the detection and calculate the metrics. In backdoor mitigation, we use an extra 1000 clean test samples to tune the model to obtain the backdoor-free model.

\noindent \textbf{Settings for Scale-Up.}
As suggested in the  \cite{guo2023scale}, we follow the same settings as the paper mentioned. We amplify the images' pixels for 1 to 11 times to get the final test datasets. And this testset is evaluated on the model and calculate the AUROC to evaluate the consistency.

\noindent \textbf{Settings for TeCo.}
Teco \cite{liu2023detecting} uses the image corruption and then evaluate the prediction results' consistency to determine whether a model is backdoored. The image corruptions we used here are gaussian noise, shot noise, impulse noise, defocus blur, motion blur, snow, frost, fog, brightness, contrast, elastic transform, pixelate and jpeg compression. The backdoored model has different prediction results on triggered images under these image corruptions.

\noindent \textbf{Settings for NAD.}
Neural Attention Distillation (NAD)  \cite{li2021neural} is a backdoor mitigation method that employs a teacher network trained on a small clean data subset to guide the fine-tuning of the backdoored student network, ensuring alignment of intermediate-layer attention. Here, we only choose the attention layer after the prompt input layer from the teacher net to instruct the learning of the student net. The reason is that in our scenario, only the parameters of the prompts are updated. We set the power of the hyper-parameter for the attention loss to 5.0 and beta to 500. The learning of the teacher network is set to 10 epochs with a learning rate of 0.01 by SGD. Moreover, the distillation process is 20 epochs with an initial learning rate of 0.01 and decay in the 4th, 8th, 12th, and 16th epochs.

\noindent \textbf{Settings for I-BAU.}
I-BAU  \cite{zeng2021adversarial} is a backdoor mitigation method that leverages implicit hyper gradient to account for the interdependence between inner and outer optimization. To solve the min-max problem in this method, we choose the Adam \cite{kingma2014adam} as our optimizer and to mitigate the influence of the I-BAU on benign accuracy, we set the learning rate to 0.0005 since the Adam has a good convergence speed and it still has a good performance on the attacks.




\begin{table}{}
\centering
\caption{The average results on VTAB-1k of TUAP and SWARM.}
\scalebox{0.8}{
\begin{tabular}{c|cc}
\toprule
Metrics  & \multicolumn{1}{c}{BA}    & ASR   \\ \hline
TUAP     & \multicolumn{1}{c}{57.27} & 92.39 \\ \hline
SWARM-B  & \multicolumn{1}{c}{\textbf{59.95}} & \textbf{97.90} \\ \bottomrule
\end{tabular}
}
\label{table:tuap}
\end{table}

\begin{table}{}
\centering
\caption{The average results on VTAB-1k of Dual-key and SWARM. Besides, P-ASR and I-ASR are the metrics to evaluate the bias problems.}
\scalebox{0.8}{
\begin{tabular}{c|ccccc}
\toprule
Metric   & \multicolumn{1}{c|}{BA}    & \multicolumn{1}{c|}{ASR}   & \multicolumn{1}{c|}{P-ASR} & \multicolumn{1}{c|}{ACC}   & I-ASR \\ \hline
Dual-key & \multicolumn{1}{c|}{43.73} & \multicolumn{1}{c|}{55.33} & \multicolumn{1}{c|}{41.13} & \multicolumn{1}{c|}{43.68} & 53.84 \\ \hline
SWARM-B  & \multicolumn{1}{c|}{\textbf{59.95}} & \multicolumn{1}{c|}{\textbf{97.90}} & \multicolumn{1}{c|}{\textbf{14.36}} & \multicolumn{1}{c|}{\textbf{61.50}} & \textbf{11.87} \\
\bottomrule
\end{tabular}
}
\label{table:Dual-key}
\end{table}

\begin{table}[]
\caption{Datasets used for backdoor defenses which are chosen from VTAB-1k. These four datasets have covered all kinds of datasets in the benchmark. They all have over 5000 test samples and the natural tasks have over 100 classes.}
\centering
\label{table:dataset}
\scalebox{0.8}{
\begin{tabular}{c|c|c|c|c|c}
\toprule
Datasets   & Description              & Classes & Train    & Val & Test   \\ \hline
CIFAR-100  & Natural & 100     & 800/1000 & 200 & 10,000 \\
Caltech101 &                          & 102     & 800/1000 & 200 & 6,084  \\ \hline
EuroSAT    & Specialized              & 10      & 800/1000 & 200 & 5,400  \\ \hline
DMLab      & Structured               & 6       & 800/1000 & 200 & 22,735 \\ \bottomrule
\end{tabular}
}
\end{table}

\begin{table*}[]
\centering
\caption{Results on the effect of the trigger learning. In each step, the learning of the trigger is indispensable since it can improve the performance both in BA and ASR. Three datasets show the correctness of our analysis.}
\label{table:ablation_appendix}
\scalebox{0.73}{
\begin{tabular}{c|c|cccc|cccc|cccc}
\toprule
/                        & Dataset & \multicolumn{4}{c|}{CIFAR100}                                                                                                                                & \multicolumn{4}{c|}{Flowers102}                                                                                                                              & \multicolumn{4}{c}{Pets}                                                                                                                                     \\ \hline
Mode                     & Metric  & \multicolumn{1}{c|}{w/o $\delta_{clean}$} & \multicolumn{1}{c|}{w/o $\delta_{backdoor}$} & \multicolumn{1}{c|}{w/o $\delta$} & w/ All & \multicolumn{1}{c|}{w/o $\delta_{clean}$} & \multicolumn{1}{c|}{w/o $\delta_{backdoor}$} & \multicolumn{1}{c|}{w/o $\delta$} & w/ All & \multicolumn{1}{c|}{w/o $\delta_{clean}$} & \multicolumn{1}{c|}{w/o $\delta_{backdoor}$} & \multicolumn{1}{c|}{w/o $\delta$} & w/ All \\ \hline
SWARM-B & BA      & \multicolumn{1}{c|}{72.33}                      & \multicolumn{1}{c|}{56.13}                         & \multicolumn{1}{c|}{28.21}                    & 76.36 & \multicolumn{1}{c|}{93.54}                      & \multicolumn{1}{c|}{70.47}                         & \multicolumn{1}{c|}{65.49}                    & 93.53 & \multicolumn{1}{c|}{80.73}                      & \multicolumn{1}{c|}{52.82}                         & \multicolumn{1}{c|}{45.68}                    & 86.02 \\ \cline{2-14} 
                         & ASR     & \multicolumn{1}{c|}{98.04}                      & \multicolumn{1}{c|}{70.92}                         & \multicolumn{1}{c|}{86.72}                    & 96.96 & \multicolumn{1}{c|}{93.44}                      & \multicolumn{1}{c|}{32.75}                         & \multicolumn{1}{c|}{60.20}                    & 96.99 & \multicolumn{1}{c|}{79.20}                      & \multicolumn{1}{c|}{53.15}                         & \multicolumn{1}{c|}{64.40}                    & 98.53 \\ \hline
SWARM-C & BA      & \multicolumn{1}{c|}{74.41}                      & \multicolumn{1}{c|}{73.87}                         & \multicolumn{1}{c|}{74.10}                    & 76.41 & \multicolumn{1}{c|}{94.35}                      & \multicolumn{1}{c|}{96.24}                         & \multicolumn{1}{c|}{97.02}                    & 96.80 & \multicolumn{1}{c|}{86.51}                      & \multicolumn{1}{c|}{86.32}                         & \multicolumn{1}{c|}{86.18}                    & 86.64 \\ \cline{2-14} 
                         & BA-T    & \multicolumn{1}{c|}{70.93}                      & \multicolumn{1}{c
                         |}{74.03}                         & \multicolumn{1}{c|}{73.99}                    & 76.38 & \multicolumn{1}{c|}{94.17}                      & \multicolumn{1}{c|}{96.50}                         & \multicolumn{1}{c|}{97.08}                    & 96.93 & \multicolumn{1}{c|}{86.15}                      & \multicolumn{1}{c|}{86.62}                         & \multicolumn{1}{c|}{86.45}                    & 86.43 \\ \bottomrule
\end{tabular}
}
\end{table*}

\section{Extra Experiments}
\subsection{Differences from TUAP and Dual-key}
In this part, we supplement two extra baseline attacks to compare their performance with our SWARM. TUAP and our SWARM use optimizing-based triggers, but TUAP only relies on the image trigger without considering the switch mechanism and has inferior performance as shown in Table \ref{table:tuap}. Bsides, Dual-key is a backdoor attack on VQA, using a fixed textual trigger, while switch token is learnable. It leads to the bias phenomenon, the textual trigger alone activates the backdoor on almost 30\% of questions as reported in paper, which makes it infeasible to apply Dual-key-like methods to achieve the switch mechanism. The results are shown in \ref{table:Dual-key}.

\subsection{Robustness to Patch Processing}
Since the Patch Processing\cite{doan2023defending} is a specific backdoor defense method designed for the vision transformers, we also evaluate SWARM's robustness to it. The results are shown in \ref{table:PP} and they indicate our method's robustness to the method specially designed for the ViTs. In summary, our method keeps high ASR-D and low AUROC under the detection which surpasses all other baseline methods.

\begin{table}{}
\centering
\caption{Patch processing defense on 6 attack methods and the results are the average of four datasets(CIFAR100, Caltech, DMLab, EuroSAT).}
\label{tab:patch_pr}
\scalebox{0.8}{
\begin{tabular}{c|cc}
\toprule
Datasets & \multicolumn{2}{c}{Average}         \\ \hline
Metrics  & \multicolumn{1}{c|}{AUROC}  & ASR-D \\ \hline
BadNets  & \multicolumn{1}{c|}{0.5669} & 37.84 \\ \hline
Blended  & \multicolumn{1}{c|}{0.5696} & \underline{44.93} \\ \hline
WaNet    & \multicolumn{1}{c|}{\textbf{0.4921}} & 37.12 \\ \hline
ISSBA    & \multicolumn{1}{c|}{0.5891} & 41.07 \\ \hline
SWARM    & \multicolumn{1}{c|}{\underline{0.5003}} & \textbf{58.48} \\ \bottomrule
\end{tabular}
}
\label{table:PP}
\end{table}

\subsection{Ablation Study on Trigger Learning}

In this part, we supplement the extra content of the ablation study which focuses on trigger learning. Although the random noise sampled from the uniform distribution can also act as a trigger, the learning method can provide a better performance both on benign accuracy and attack success rate which is very important in our method.

As we can see in \cref{table:ablation_appendix}, without trigger learning in clean mode, the visual prompts have an obvious accuracy drop especially in the triggered images in the clean mode while ASR has no performance decrease. In contrast, without trigger learning in backdoor mode, both BA and ASR suffer a big drop in backdoor mode. And finally, if we keep the random noise as the trigger, the backdoor attack can not be established successfully since the BA in backdoor mode is very low.

All the experiments on three datasets have shown the trigger learning's importance in our method. The trigger learning in two modes has balanced the performance on benign accuracy and the attack success rate.

In all, each component in our method has been analyzed and shows its indispensability in our method.

\subsection{Effect of $\epsilon$}
\begin{figure}[]
    \centering
    \includegraphics[width=0.46\textwidth]{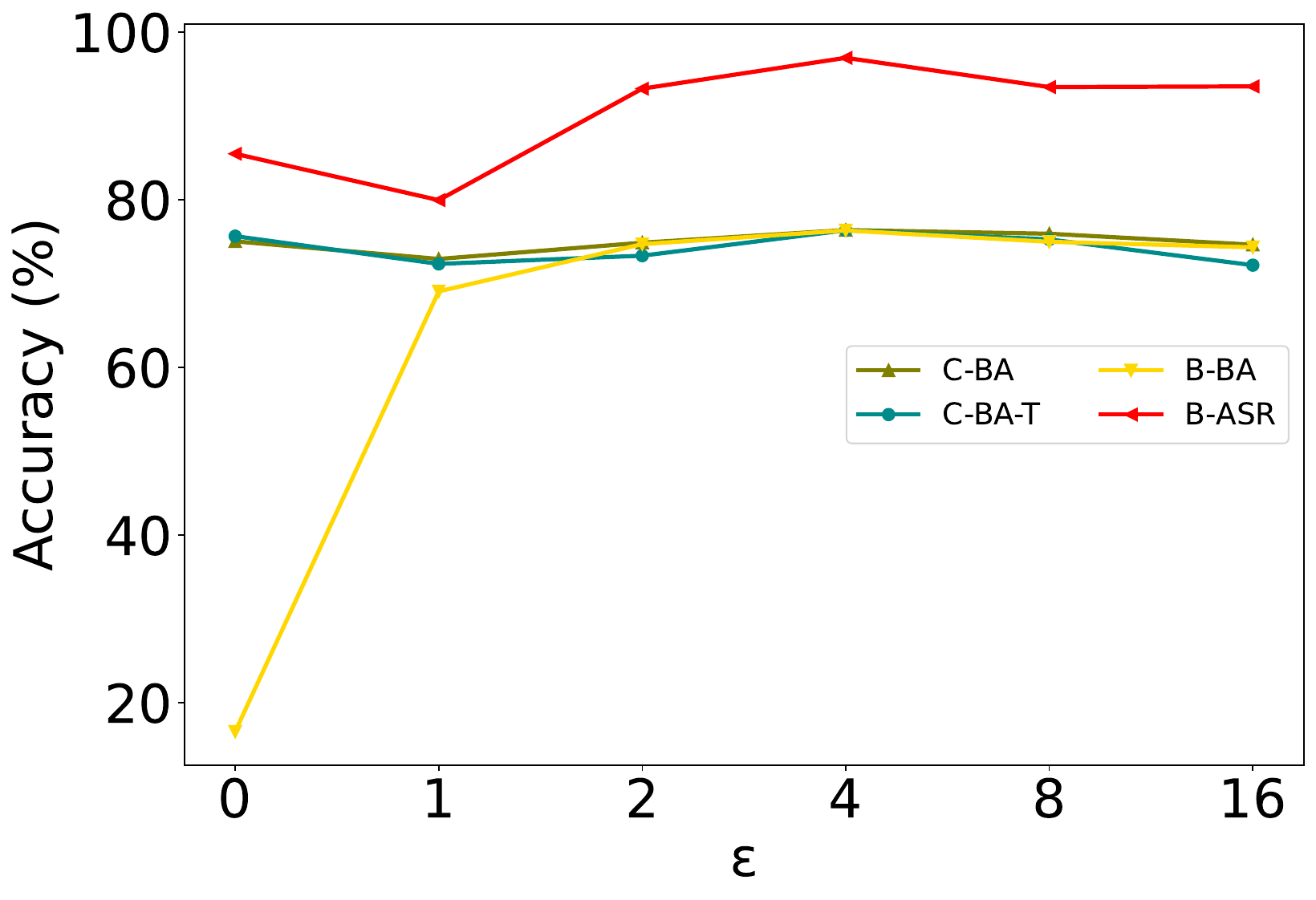}
        \caption{The effect of increasing the $\epsilon$. If $\epsilon$ is 0, the benign accuracy in the backdoor mode is very poor.}
    \label{fig:epsilon}
\end{figure}
As is shown in \cref{fig:epsilon}, we evaluate the effect of the $\epsilon$ on CIFAR100. The $\epsilon$ is the noise limit implemented on the trigger. The $l^{\infty}$ restriction is used here so $||\delta||_{\infty}\le \epsilon$. When the $\epsilon=0$, it means that we don't adopt the trigger in our method. In \cref{fig:epsilon}, $\epsilon=0$ makes the benign accuracy drop a lot. With the increase of the $\epsilon$, the performance on BA and ASR both in clean mode and backdoor mode has improved and achieved the peak when $\epsilon=4$. And the performance keep stable with the $\epsilon$ goes on increasing.

\subsection{Effect of Prompt Length}
\begin{figure}[]
    \centering
    \includegraphics[width=0.46\textwidth]{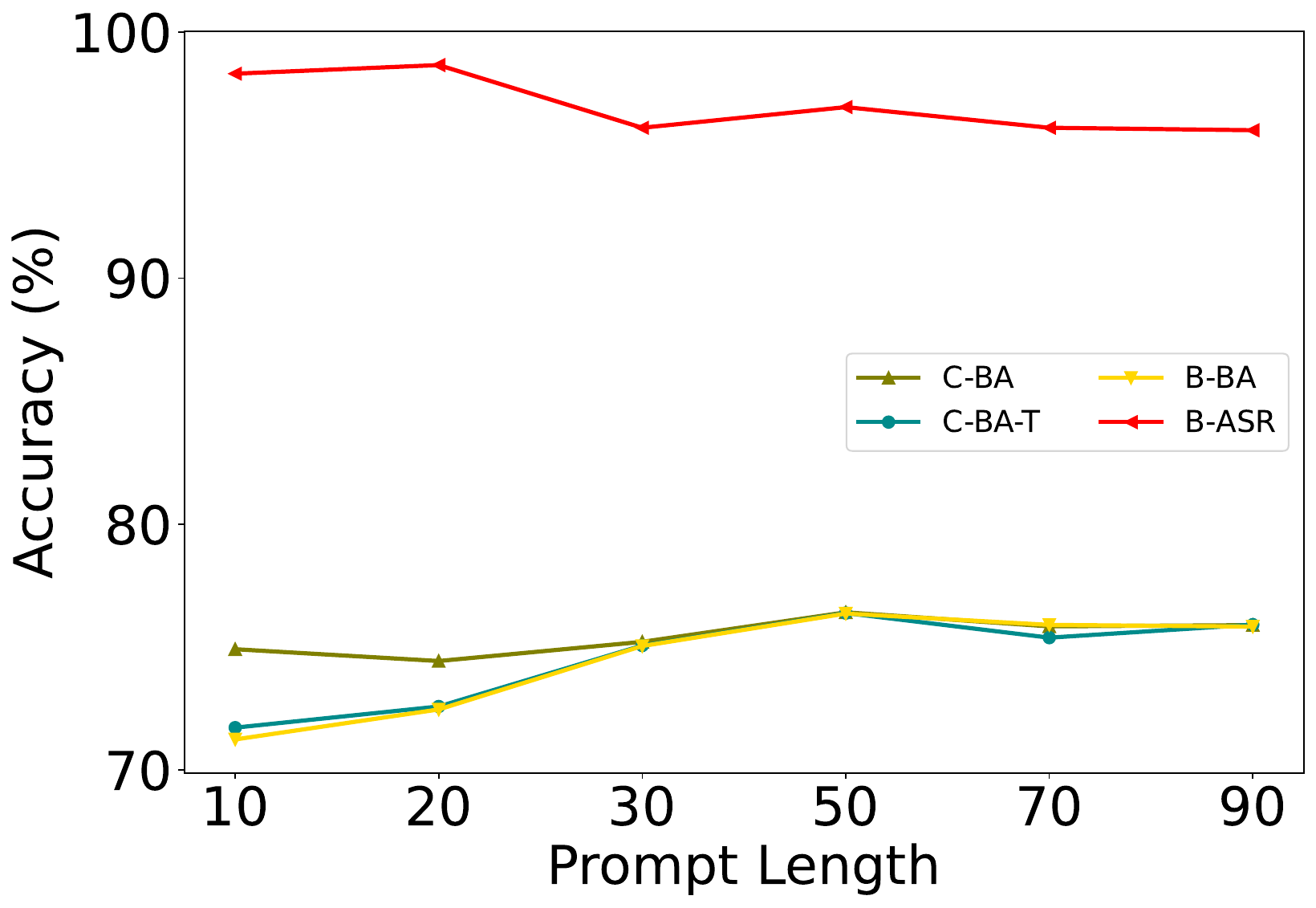}
        \caption{The effect of increasing the prompt length. SWARM has a stable performance when the prompt length varies.}
    \label{fig:length}
\end{figure}
As is shown in the \cref{fig:length}, the experiments are done on CIFAR100. Even when the prompt length has a big variance, our method still has a stable effect on attacking the model. When the prompt length is 10, the triggered images have a 4\% drop compared to the peak in BA-T in clean mode. The BA in backdoor mode also suffers a drop. However, when the prompt length increases, the performance on these two metrics has a huge improvement and gradually achieves the peak when the prompt length is 50. In contrast, ASR in backdoor mode still keeps over 95\% in all lengths of prompts and the clean images have the same performance when the prompt length decreases.

\subsection{Robustness to STRIP}
\begin{table*}[]
\caption{The defense results on Fine-tuning. Our method still keeps
high ASRs after the mitigation comparing to other baselines.}
\label{table:Finetuning}
\centering
\scalebox{0.8}{
\begin{tabular}{c|cc|cc|cc|cc|cc}
\toprule
Attack         & \multicolumn{2}{c|}{BadNets}       & \multicolumn{2}{c|}{Blended}       & \multicolumn{2}{c|}{WaNet}         & \multicolumn{2}{c|}{ISSBA}         & \multicolumn{2}{c}{SWARM}          \\ \hline
Dataset,Metric & \multicolumn{1}{c|}{BA}    & ASR   & \multicolumn{1}{c|}{BA}    & ASR   & \multicolumn{1}{c|}{BA}    & ASR   & \multicolumn{1}{c|}{BA}    & ASR   & \multicolumn{1}{c|}{BA}    & ASR   \\ \hline
CIFAR100       & \multicolumn{1}{c|}{64.41} & 86.06 & \multicolumn{1}{c|}{63.51} & 85.04 & \multicolumn{1}{c|}{47.63} & 42.85 & \multicolumn{1}{c|}{73.00}    & 11.61 & \multicolumn{1}{c|}{76.52} & 96.97 \\ \hline
Caltech101     & \multicolumn{1}{c|}{61.38} & 35.22 & \multicolumn{1}{c|}{58.62} & 33.64 & \multicolumn{1}{c|}{66.23} & 29.84 & \multicolumn{1}{c|}{64.68} & 34.49 & \multicolumn{1}{c|}{78.19} & 95.97 \\ \hline
EuroSAT        & \multicolumn{1}{c|}{89.16} & 95.94 & \multicolumn{1}{c|}{90.55} & 97.02 & \multicolumn{1}{c|}{77.02} & 28.39 & \multicolumn{1}{c|}{91.50}  & 18.57 & \multicolumn{1}{c|}{90.43} & 96.05 \\ \hline
DMLab          & \multicolumn{1}{c|}{33.79} & 97.59 & \multicolumn{1}{c|}{34.98} & 99.55 & \multicolumn{1}{c|}{34.51} & 80.18 & \multicolumn{1}{c|}{35.44} & 37.07 & \multicolumn{1}{c|}{32.83} & 96.30  \\ \bottomrule
\end{tabular}
}
\end{table*}

\begin{figure}[]
    \centering
    \includegraphics[width=0.46\textwidth]{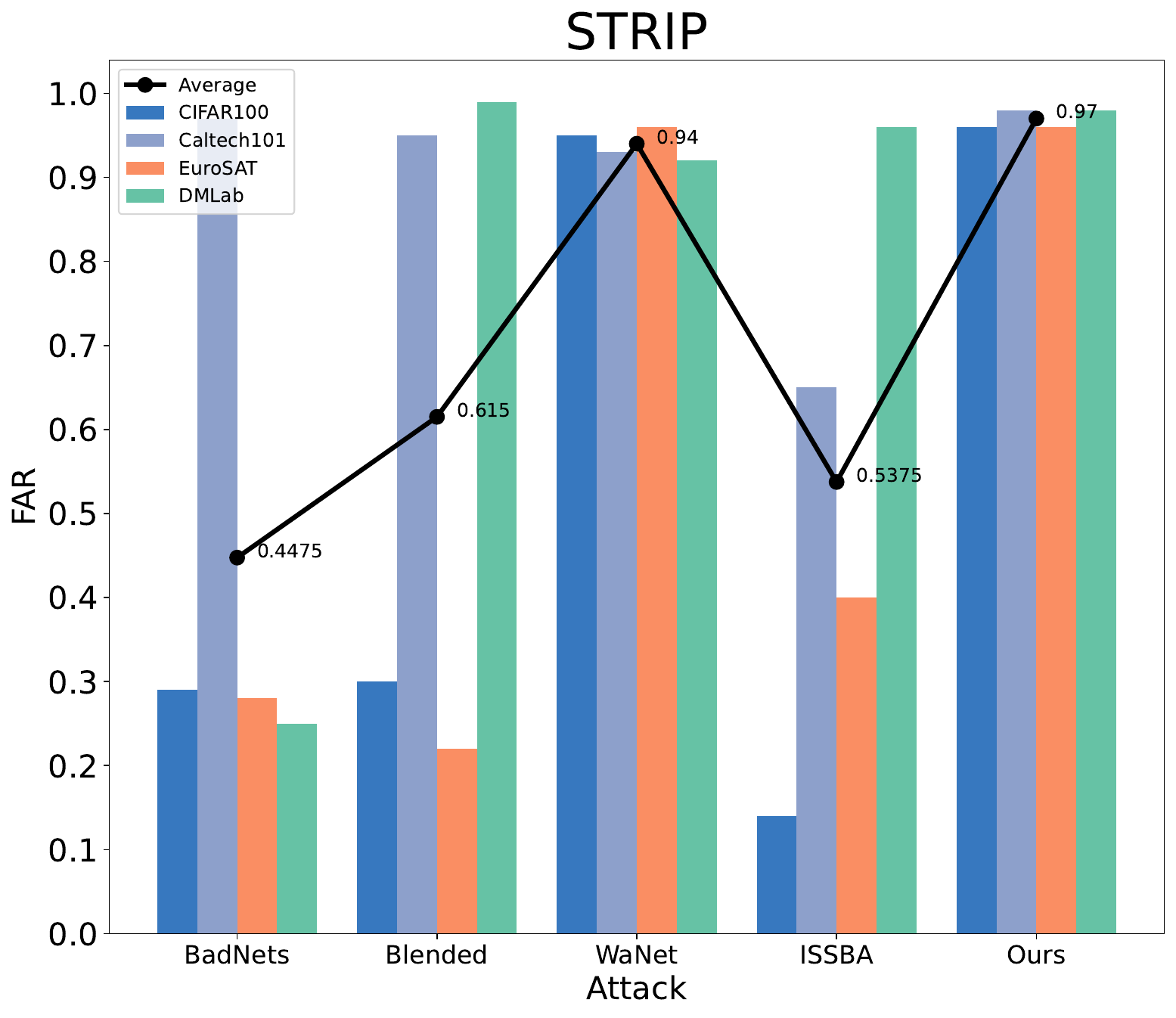}
        \caption{The results of STRIP detection methods on five backdoor attacks. Higher FAR indicates a better attack performance. Among these attacks, SWARM exceeds all other baseline attacks.}
    \label{fig:STRIP}
\end{figure}

\begin{figure*}[]
    \centering
    \includegraphics[width=0.98\textwidth]{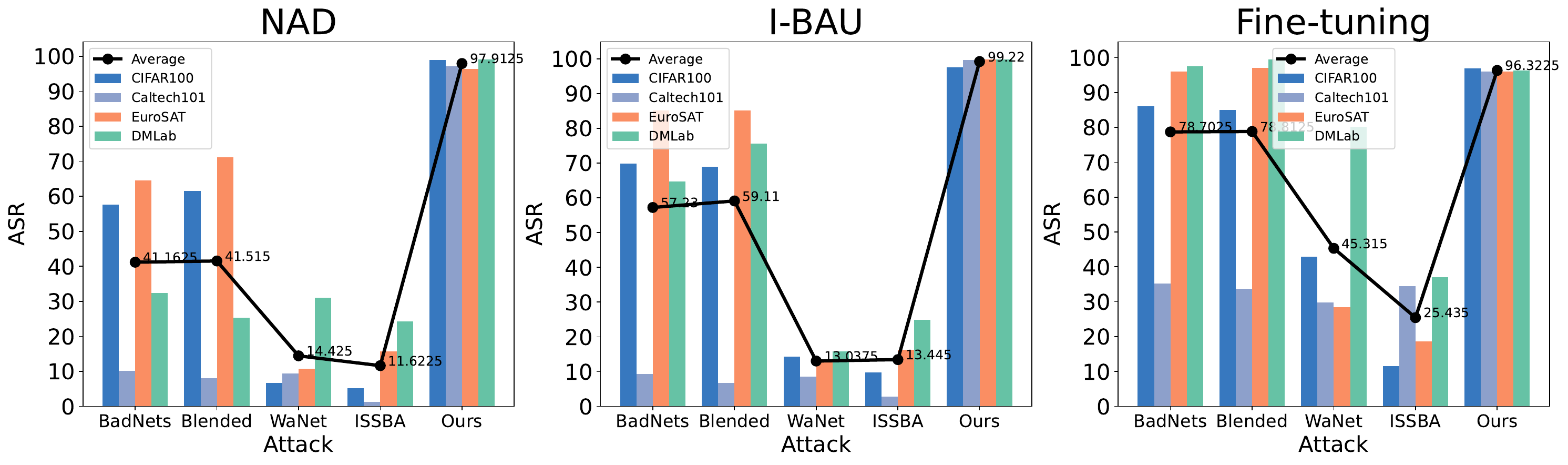}
        \caption{The results of three backdoor mitigation methods against five backdoor attacks on four datasets. ASR is the metric shown in the figure and SWARM has shown over 95\% ASR in every situation. Our method is superior to all other baseline attacks.}
    \label{fig:mitigations}
\end{figure*}
\begin{figure*}[]
    \centering
    \includegraphics[width=0.98\textwidth]{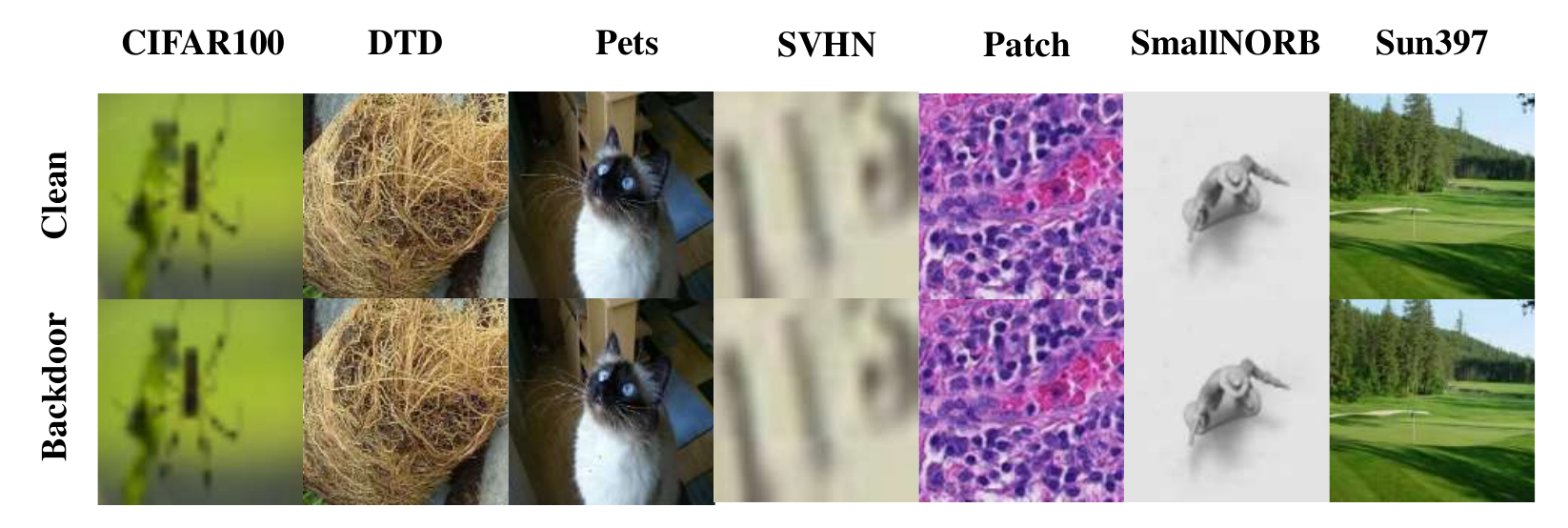}
        \caption{The visualization of datasets in VTAB-1k. As we can see, the triggered images is imperceptible from the clean images by eyes.}
    \label{fig:vis}
\end{figure*}

STRIP \cite{gao2019strip} is a classical method proposed to detect the backdoor in the given model. It intentionally perturbs the incoming input by superimposing various image patterns. Then it observes the randomness of predicted classes for perturbed inputs to determine whether the given deployed model is malicious or benign. As a result, a low entropy in predicted classes violates the input dependence property of a benign model and this phenomenon implies the presence of a malicious image.


Different from the methods proposed in the main paper whose main metric is AUROC, STRIP exploits FRR and FAR as the main metrics to evaluate the detection results. FRR is the false rejection rate which is the probability when the benign images are thought of as triggered images by the STRIP detection system. In contrast, FAR is the false acceptance rate which is the probability that the triggered image is recognized as a benign image by the STRIP detection system. The smaller FAR means the better detection effect.

Following the same settings, we use 3000 clean images and 3000 triggered images to do the test on four different datasets and we calculate the average scores to better evaluate the performance of each baseline. 

As is shown in \cref{fig:STRIP}, SWARM has shown the perfect ability to avoid STRIP's detection since all SWARM's FARs are over 0.95 on four different datasets. In \cref{fig:STRIP}, we set FRR to 0.05 to keep the same as the original paper. However, BadNets, Blended, and WaNet don't have such a high performance on average since they have almost 0.5 FAR under the detection which is much lower than SWARM's FARs. On the other hand, WanNet has a high FAR but it has already been proved that it has poor performance in ASR-D which is also a not successful attack. We can get the same conclusion as in the main paper, SWARM has surpassed all other baseline attacks in resisting the detection method.

\subsection{Average ASR under Backdoor Mitigations}

\subsection{Robustness to Fine-tuning}
Fine-tuning is a widely used method to adapt the model to the downstream tasks' domains which exploits a small amount of test samples to tune the model. Moreover, it also has the effect of mitigating the backdoors in the malicious models \cite{liu2018fine}. Therefore, we evaluate the fine-tuning's effect on our method and baseline attacks on Vision Transformers. Here, we use the extra 1000 clean test images to do the tuning and use BA and ASR to evaluate the fine-tuning's mitigation effect. The learning rate is 0.001 which is small enough so as not to destroy the parameters learned by the training process. Besides, the learning epoch is also set to 10 to avoid the same problem.

As is shown in \cref{table:Finetuning}, we have done the experiments on four different datasets. SWARM still has the best performance to resist the influence of backdoor mitigation. The SWARM's ASRs are all kept over 95\%. However, other baseline attacks all have failed cases in different datasets which shows our method's robustness. In all, ISSBA has the poorest performance on resisting the backdoor mitigation since its main contribution is to resist the detection methods. In Caltech101 \cite{fei2006one}, all the baseline attacks have low ASRs on this dataset. We hypothesize this dataset is the most difficult task in these four datasets so the backdoors are easy to be mitigate but SWARM still keeps a high ASR on this dataset.

As is shown in \cref{fig:mitigations}, we compare five backdoor attacks on four datasets with three mitigation methods. It's noteful that our SWARM surpasses all the baseline attacks in resisting the mitigation methods and has over 95\% ASRs in all of these situations. The average results have the same trend on different mitigation methods and NAD shows the best mitigation performance on decreasing ASRs. However, ISSBA has the worst performance on resisting the backdoor mitigations.

\subsection{More Visualizations on Triggered Images}
As is shown in \cref{fig:vis}, we have exhibited the clean images and triggered images from 7 datasets. The triggers in these images are imperceptible.

\section{Social Impact}
In all, we have proposed a switchable backdoor attack that is difficult to detect and remove. Such a kind of attack can exist in the pre-trained models' adapting process which introduces a small amount of learnable parameters to fit for the downstream tasks. This kind of attack is practical to happen in the real world if the cloud service is an adversary and this kind of attack is more dangerous since it can resist the state-of-the-art backdoor defenses including detections and backdoor mitigations.

We have considered the possible huge impact of proposing such a backdoor attack in the pre-training era. This kind of backdoor attack is easy to implement, resource-efficient, and hard to detect and mitigate. The adapting process can exploit this method to provide a malicious service which may cause huge harm to the whole society.

However, as the era of the pre-training model emerges, the security and trustworthiness of this paradigm need more attention. Therefore, we propose such a backdoor attack so as to hope the community to pay more attention to such a two-mode attack during the adaption and in the future to build a reliable enough machine learning system to service the whole society. We hope the pre-training paradigm can further improve human life.


\end{document}